%% file: main.tex
\documentclass[sigconf,authorversion,nonacm]{acmart}

\AtBeginDocument{%
  }

\setcopyright{acmlicensed}
\copyrightyear{2025}
\acmYear{2025}
\acmDOI{XXXXXXX.XXXXXXX}
\acmConference[ACM KDD '25]{Make sure to enter the correct
  conference title from your rights confirmation email}{August 03--07,
  2025}{Toronto, ON}
\acmISBN{978-1-4503-XXXX-X/2018/06}

\usepackage{microtype}
\usepackage{graphicx}
\usepackage{subfigure}
\usepackage{subcaption}
\usepackage{booktabs} 

\usepackage{hyperref}
\usepackage{graphicx}
\usepackage{caption} 
\usepackage{ragged2e}
\usepackage{algorithm}
\usepackage{algorithmic}

\usepackage{amsmath}
\usepackage{mathtools}
\usepackage{amsthm}

\usepackage[capitalize,noabbrev]{cleveref}

\theoremstyle{plain}
\newtheorem{theorem}{Theorem}[section]
\newtheorem{proposition}[theorem]{Proposition}

\theoremstyle{definition}
\newtheorem{definition}[theorem]{Definition}

\theoremstyle{remark}

\newtheorem*{remark*}{Remark}

\usepackage{url}
\usepackage{array} 
\usepackage{comment}
\usepackage{multirow} 
\usepackage{siunitx}
\usepackage[utf8]{inputenc} 
\usepackage[T1]{fontenc}    
\usepackage{amsfonts}       
\usepackage{nicefrac}       
\usepackage{enumitem}
\usepackage{dsfont}
\usepackage{tcolorbox}

\begin{document}

\title{\textit{Ordered} Topological Deep Learning: a Network Modeling Case Study}


\author[Guillermo Bernárdez, Miquel Ferriol-Galmés, et al.]{Guillermo Bernárdez$^{*}$\textsuperscript{1}, Miquel Ferriol-Galmés$^{*}$\textsuperscript{2}, Carlos Güemes-Palau\textsuperscript{2}, Mathilde Papillon\textsuperscript{1}, \\Pere Barlet-Ros\textsuperscript{2}, Albert Cabellos-Aparicio\textsuperscript{2}, Nina Miolane\textsuperscript{1}}
\authornote{$^{*}$Both authors contributed equally to this research and are designated as corresponding authors. For theory-related questions, please contact Guillermo Bernárdez \textless{}guillermo\_bernardez@ucsb.edu\textgreater{}; for experimental results, please contact Miquel Ferriol-Galmés \textless{}miquel.ferriol@upc.edu\textgreater{}.}
\affiliation{%
  \institution{\textsuperscript{1}UC Santa Barbara, USA \textsuperscript{2}Universitat Politecnica de Catalunya, Spain}
  \country{}
}



\begin{abstract}
Computer networks are the foundation of modern digital infrastructure, facilitating global communication and data exchange. As demand for reliable high-bandwidth connectivity grows, advanced network modeling techniques become increasingly essential to optimize performance and predict network behavior. Traditional modeling methods, such as packet-level simulators and queueing theory, have notable limitations --either being computationally expensive or relying on restrictive assumptions that reduce accuracy. In this context, the deep learning-based RouteNet family of models has recently redefined network modeling by showing an unprecedented cost-performance trade-off. 
In this work, we revisit RouteNet's sophisticated design and uncover its hidden connection to Topological Deep Learning (TDL), an emerging field that models higher-order interactions beyond standard graph-based methods. We demonstrate that, although originally formulated as a heterogeneous Graph Neural Network, RouteNet serves as the first instantiation of a new form of TDL.
More specifically, this paper presents OrdGCCN, a novel TDL framework that introduces the notion of \textit{ordered neighbors} in arbitrary discrete topological spaces, and shows that RouteNet's architecture can be naturally described as an ordered topological neural network.
To the best of our knowledge, this marks the first successful real-world application of state-of-the-art TDL principles --which we confirm through extensive testbed experiments--, laying the foundation for the next generation of ordered TDL-driven applications.  
\end{abstract}






\settopmatter{printfolios=true}
\maketitle

\section{Introduction}
\label{sec:intro}
\input{sections/01-introduction}

\section{Topological Deep Learning}
\label{sec:tdl}
\input{sections/02-tdl}

\section{Computer Networks: a New Perspective}
\label{sec:computer_networks}
\input{sections/03-computer_networks}

\section{Introducing Order in TDL}
\label{sec:order_TDL}
\input{sections/04-order_TDL}

\section{RouteNet: Network Modeling Meets TDL}
\label{sec:routenet}
\input{sections/05-routenet}

\section{(Topo)RouteNet in Action}
\label{sec:experiments}
\input{sections/06-experiments}
\section{Concluding Remarks}
\label{sec:implications}
\input{sections/07-implications}
\bibliographystyle{ACM-Reference-Format}
\bibliography{bibliography}

\clearpage
\appendix
\input{sections/appendices/proofs}

\section{Background on Network Modeling}
\label{app:networking_background}

\input{sections/appendices/networking_background}

\section{Evolution of RouteNet}
\label{app:evolution_of_routenet}
\input{sections/appendices/evolution_of_routenet}

\section{Technical Details on RouteNet's Architecture}
\label{app:routenet_technical_details}
\input{sections/appendices/routenet_technical_details}

\section{Nature and Characteristics of Network Data Generation}
\label{app:network_data}
\input{sections/appendices/network_data_details}

\section{Summary of Experimentation of RouteNet models}
\label{app:routenet_experiments}
\input{sections/appendices/routenet_experiments}

\section{Additional Benchmark Results}
\label{app:benchmarks}
\input{sections/appendices/additional_benchmarks}

\end{document}

%% file: sections/01-introduction.tex
In today's interconnected world, computer networks form the backbone of our digital infrastructure, enabling seamless communication and data exchange across the globe. From personal communication and social media to critical services such as healthcare, finance, and transportation, virtually every aspect of modern life relies on robust and efficient computer networks. As the need for faster data speeds and more reliable connectivity increases, so does the need for advanced technologies to manage and optimize these complex networks. The ability to accurately model and predict network behavior is crucial for ensuring that networks can meet the evolving needs of society, making network modeling a vital area of research and development.

\begin{figure*}[!htbp]
    \centering
    \begin{minipage}{0.48\textwidth}
        \flushleft  
        \textbf{A.}\\
        \centering
        \includegraphics[width=\textwidth]{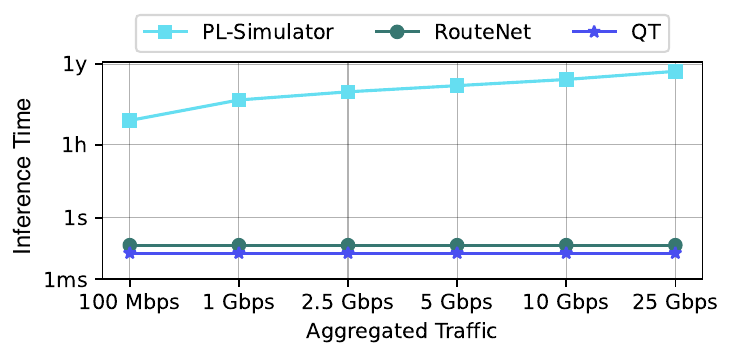}
        \Description{A graph showing inference times for simulating 1 second of network operation with varying traffic levels.}
    \end{minipage}
    \hfill
    \begin{minipage}{0.48\textwidth}
        \flushleft  
        \textbf{B.}\\
        \centering
        \includegraphics[width=\textwidth]{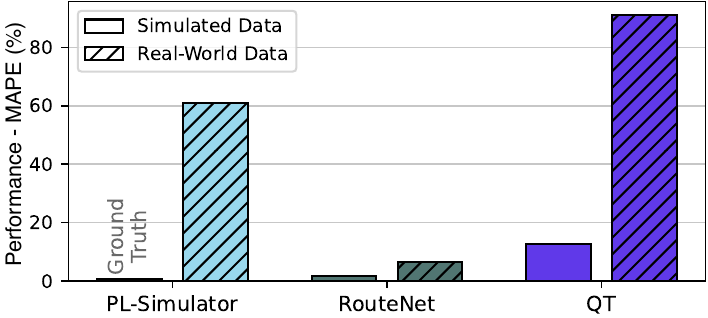}
        \Description{A bar chart comparing delay prediction performance (MAPE) for different methods using simulated and real traffic data.}
    \end{minipage}

    \caption{Comparison of RouteNet against traditional network modeling. A. Inference times for simulating 1 second of network operation in a fixed topology depending on the amount of traffic found in the network. B. Delay prediction performance (Mean Absolute Percentage Error (MAPE), lower is better) obtained by each of the methods with both simulated and real traffic data.}
    \label{fig:comparison_sota}
\end{figure*}

Network modeling is a fundamental tool in the networking community, providing the means to simulate, analyze, and optimize the performance of network systems. Accurate models allow network engineers to predict traffic patterns, identify potential bottlenecks, and design more efficient routing protocols. Traditional network modeling methods have relied heavily on packet-level simulators \cite{varga2001discrete, riley2010ns} and analytical models such as Queueing Theory (QT)~\cite{czachorski2014queueing}. 
By simulating the movement of each individual packet as it traverses the network, packet-level simulators ensure a high degree of accuracy, but at the expense of being computationally expensive and impractical for large-scale networks. On the other hand, QT provides a faster alternative but is limited by its simplifying assumptions --such as Poisson distribution for arrival times--, which can lead to inaccuracies in real-world scenarios.


To overcome the limitations of traditional network modeling methods, researchers have leveraged deep learning techniques to develop more accurate and scalable network models. In recent years, several deep learning-based approaches have emerged as promising alternatives \cite{zhang2021mimicnet, yang2022deepqueuenet}, capable of learning complex patterns in network data and making precise predictions without the need for restrictive assumptions.
Among these innovative approaches, the RouteNet family of models \cite{rusek2020routenet,ferriol2022routenet,ferriol2023routenet,guemes2025routenet} has shown significant potential in addressing the shortcomings of traditional network modeling techniques. 

Originally devised as a heterogeneous Graph Neural Network (GNN)~\cite{scarselli2008graph}, RouteNet is able to model network behavior more accurately and efficiently. By capturing the intricate relationships between network components through a multiple-step message-passing scheme, RouteNet can predict performance metrics with high precision, even in large and complex network environments. This breakthrough represents a significant advancement in the field, offering a viable path toward more reliable and scalable network modeling solutions.

This work uncovers the theoretical principles that drive RouteNet's success. To this aim, we leverage a key insight: even though RouteNet was designed with GNNs, it builds an internal representation of the network that goes beyond the pairwise connections and local neighborhoods of the graph domain. In fact, by examining how RouteNet actually models multi-element relationships, we uncover its strong connection to Topological Deep Learning (TDL)~\cite{bodnar2023thesis,hajij2023tdl} -- an emerging field that offers a principled framework to encode higher-order interactions. TDL is widely regarded as a promising framework for modeling complex systems -- such as particle physics, social interactions, or biological networks \citep{lambiotte2019networks}. To date, however, a key challenge remains: the lack of practical use cases that convincingly demonstrate its real-world effectiveness~\citep{papamarkou2024position}. Here, we establish a theoretical connection between TDL and RouteNet by introducing an extension of TDL: ordered TDL. By showing that RouteNet represents an instantiation of ordered TDL, we provide the first compelling example of how TDL can be successfully applied to a real-world problem, highlighting its potential to revolutionize network modeling and beyond.

\vspace{10pt}
\paragraph{Contributions.} This paper revisits the RouteNet family of architectures through the lens of TDL, revealing the topological transformations that these models have implicitly employed in their design. 
More specifically, our key contributions are:
\begin{itemize}
    \item We show that computer networks naturally admit higher-order topological representations.
    
    \item We propose a novel TDL framework - ordered TDL - that introduces the general notion of order in arbitrary discrete, higher-order topological spaces. We introduce its neural network architectures called Ordered Generalized Combinatorial Complex Networks (OrdGCCNs). 

    \item We prove that, by enabling aggregators that are not permutation invariant, OrdGCCNs become the most expressive Topological Neural Network (TNN) to date.
    
    \item We show that RouteNet, the top reference ML family of models in the network modeling field, can naturally be rewritten as a OrdGCCN. To the best of our knowledge, this represents the first cutting-edge TDL-based application to a real-world setting.
    
    \item We conduct new simulation and testbed experiments that further validate OrdGCCN's state-of-the-art effectiveness in network modeling, demonstrating the superiority of RouteNet over traditional NN and GNN architectures.

\end{itemize}
Together, these contributions establish the theoretical and empirical foundations for ordered TDL, bridge the gap between TDL and real-world impact via the domain of computer networks, and pave the way for future advancements beyond network modeling.

%% file: sections/02-tdl.tex
The emerging field of Topological Deep Learning (TDL) 
aims to go beyond Graph Neural Networks (GNNs) by naturally processing relationships between more than two elements at once, which are ubiquitous in any realistic complex system (e.g., social interactions within a community, molecular structures and reactions, n-body interactions). This section introduces key TDL concepts relevant to this work. 

\begin{figure}[!t]
    \centering
    \includegraphics[width=\columnwidth]{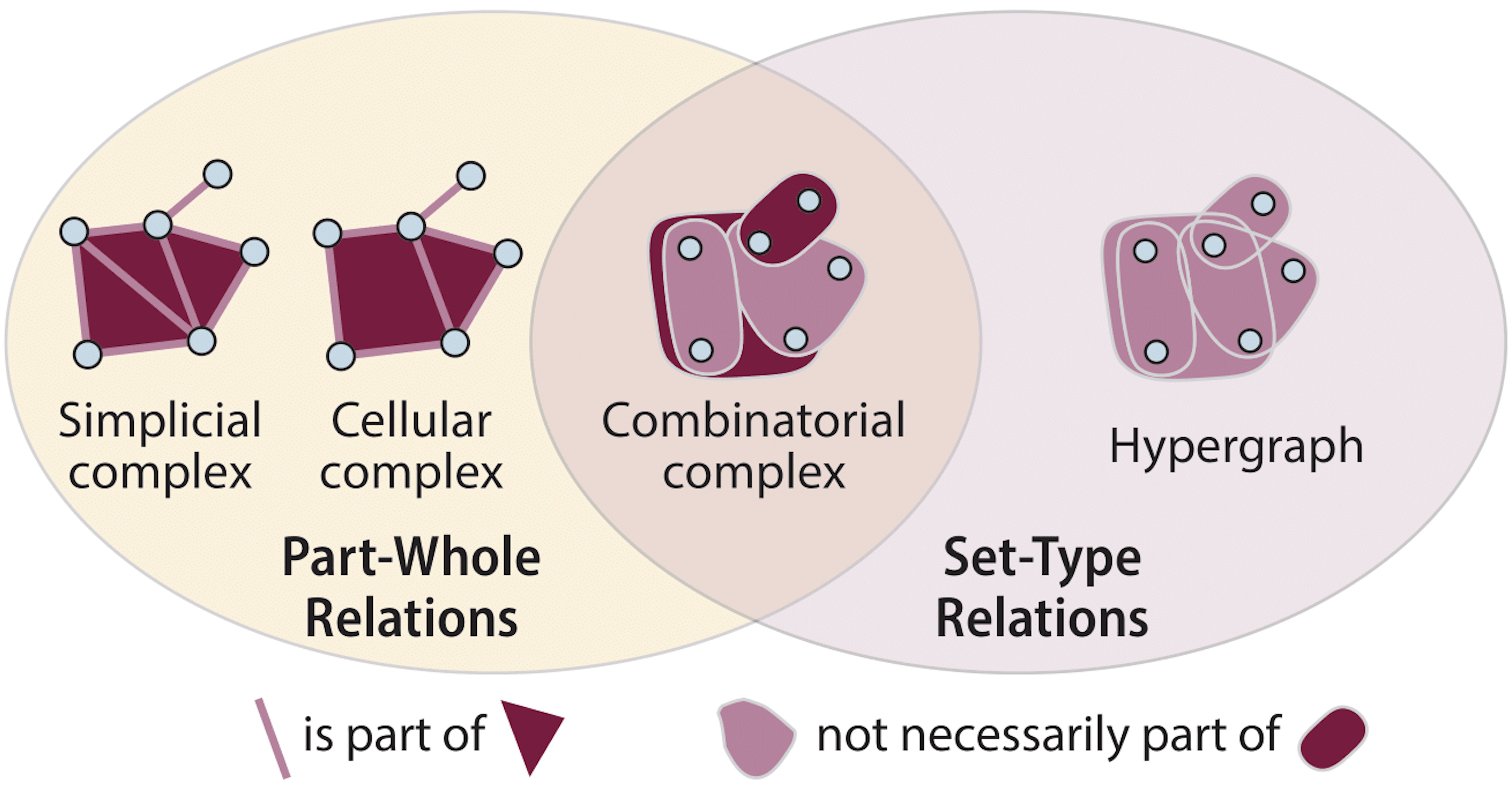}
    \caption{Domains of Topological Deep Learning encoding higher-order relations (cells in light and dark pink) between elements (nodes in blue). Figure adapted from \cite{papillon2023architectures}.}
    \label{fig:domains}
\end{figure}

\paragraph{TDL Domains.} TDL methods expand graphs' pairwise connections by encoding higher-order relationships through algebraic topology concepts. Fig.~\ref{fig:domains} illustrates the standard discrete, higher-order topological spaces used to that end, enabling more complex relational representations via part-whole and set-types relations~\cite{papillon2023architectures}. For the sake of generality, this work focuses on combinatorial complexes, which subsume all other discrete topological domains \cite{hajij2023tdl}.

\paragraph{Combinatorial Complex.} A \emph{combinatorial complex} is a triple $(\mathcal{V}, \mathcal{C}, $ $\textrm{rk})$ consisting of a set $\mathcal{V}$, a subset $\mathcal{C}$ of the power set $\mathcal{P}(\mathcal{V}) \backslash\{\emptyset\}$, and a rank function $\textrm{rk}: \mathcal{C} \rightarrow \mathbb{Z}_{\geq 0}$ with the following properties:
\begin{enumerate}[leftmargin=*,align=left,widest=3]
    \item for all $v \in \mathcal{V},\{v\} \in \mathcal{C}$ and $\textrm{rk}(\{v\})=0$;
    \item the function $\textrm{rk}$ is order-preserving, i.e., if $\sigma, \tau \in \mathcal{C}$ satisfy $\sigma \subseteq \tau$, then $\textrm{rk}(\sigma) \leq$ $\textrm{rk}(\tau)$;
\end{enumerate}
The elements of $\mathcal{V}$ represent the nodes, while the elements of $\mathcal{C}$ are called cells (i.e., groups of nodes). 
The rank of a cell $\sigma \in \mathcal{C}$ is $k:=\textrm{rk}(\sigma)$, and we call it a $k$-cell. $\mathcal{C}$ simplifies notation for $(\mathcal{V}, \mathcal{C}, \textrm{rk})$, and its dimension is defined as the maximal rank among its cells: $\mathrm{dim}(\mathcal{C}):= \max_{\sigma \in \mathcal{C}} \textrm{rk}(\sigma)$.

\begin{figure}
    \centering
    \includegraphics[width=\linewidth]{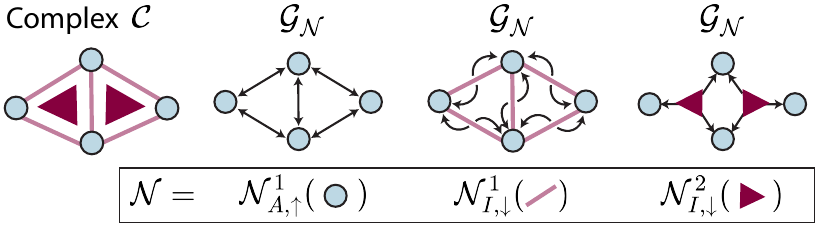}
    \caption{Neighborhoods. Given a complex $\mathcal{C}$ (left), we illustrate three examples of augmented Hasse graphs $\mathcal{G}_\mathcal{N}$ corresponding to a given neighborhood $\mathcal{N}$, listed at the bottom.}
    \label{fig:neighborhoods}
\end{figure}

\paragraph{Neighborhoods and Augmented Hasse Graphs.}
Combinatorial complexes are equipped with a notion of neighborhood among cells that confers on them a topological structure. In particular, a neighborhood $\mathcal{N}: \mathcal{C} \rightarrow \mathcal{P}(\mathcal{C})$ on a combinatorial complex $\mathcal{C}$ is a function that assigns to each cell $\sigma$ in $\mathcal{C}$, a collection of ``neighbor cells" $\mathcal{N}(\sigma) \subset \mathcal{C}\cup \emptyset$.  Examples of neighborhood functions are incidences (connecting cells with different ranks) and adjacencies (connecting cells with the same rank), although other neighborhoods can be defined for specific tasks \citep{battiloro2024n}. Additionally, each particular neighborhood $\mathcal{N}$ induces a strictly augmented Hasse graph $\mathcal{G}_{\mathcal{N}} = (\mathcal{C}_\mathcal{N}, \mathcal{E}_\mathcal{N})$ on $\mathcal{C}$ \citep{papillon2024topotune,hajij2023tdl}, defined as the directed graph whose nodes and edges are given, respectively, by $\mathcal{C}_\mathcal{N} =\{\sigma \in \mathcal{C} \, | \, \mathcal{N}(\sigma) \neq \emptyset \}$ and $\mathcal{E}_\mathcal{N} =\{(\tau, \sigma) \, | \, \tau \in \mathcal{N}(\sigma)\}$. We provide examples in Fig.~\ref{fig:neighborhoods}.

\paragraph{Faces, Cofaces, and Adjacencies.} 
In general, the $r$-up/down incidence neighborhoods of a cell $\sigma\in\mathcal{C}$ are respectively defined as 
\begin{equation}\label{eq:incidences}
\begin{split}
    \mathcal{N}_{I,\uparrow}^r(\sigma) = \{\tau \in \mathcal{C} \, | \, \textrm{rk}(\tau) = \textrm{rk}(\sigma)+r, \sigma \subset \tau\}, \\
     \mathcal{N}_{I,\downarrow}^r(\sigma) =\{\tau \in \mathcal{C} \, | \, \textrm{rk}(\tau) = \textrm{rk}(\sigma)-r, \tau \subset \sigma\},
\end{split}
\end{equation}
with $r\in \mathbb{Z}_{\geq 0}$. Considering $\textrm{rk}(\sigma)=k$, we refer to $\mathcal{N}_{I,\uparrow}^r(\sigma)$ as the set of $(k+r)$-cofaces of $\sigma$, and to $\mathcal{N}_{I,\downarrow}^r(\sigma)$ as the set of its $(k-r)$-faces. Therefore, a cell $\tau\in\mathcal{C}$ is a $(k+r)$-coface of $\sigma$ if it contains $\sigma$ and $\textrm{rk}(\tau) = k+r$; analogously, $\tau\in\mathcal{C}$ is called a $(k-r)$-face of $\sigma$ if is is contained by $\sigma$ and $\textrm{rk}(\tau) = k-r$. These incidence neighborhoods induce $r$-up/down adjacencies as
\begin{equation}\label{eq:adjacencies}
\begin{split}
     \mathcal{N}_{A,\uparrow}^r(\sigma) = \{\tau \in \mathcal{C} \, | \, \textrm{rk}(\tau)=\textrm{rk}(\sigma), \mathcal{N}_{I,\uparrow}^r(\sigma) \cap \mathcal{N}_{I,\uparrow}^r(\tau) \neq \emptyset  \}, \\ 
     \mathcal{N}_{A,\downarrow}^r(\sigma) = \{\tau \in \mathcal{C} \, | \, \textrm{rk}(\tau)=\textrm{rk}(\sigma), \mathcal{N}_{I,\downarrow}^r(\sigma) \cap \mathcal{N}_{I,\downarrow}^r(\tau) \neq \emptyset \}.
\end{split}
\end{equation}
Therefore, two $k$-cells $\sigma$ and $\tau$ are said to be $(k+r)$-adjacent if they are both contained in a $(k+r)$-cell $\delta\in\mathcal{C}$; analogously, they are called $(k-r)$-adjacent if they both contain a $(k-r)$-cell $\delta\in\mathcal{C}$. 

\paragraph{Combinatorial Complex Signals.} A signal over a combinatorial complex $\mathcal{C}$ is a mapping $f: \mathcal{C} \rightarrow \mathbb{R}^F$ from the set of cells $\mathcal{C}$ to feature space $\mathbb{R}^F$. In particular, the feature vector $h_\sigma$  of a cell $\sigma \in \mathcal{C}$ is typically defined as $h_\sigma = [f_1(\sigma),\dots,f_F(\sigma)]\in \mathbb{R}^F$. 

\paragraph{Topological Neural Networks.}
The work of~\citet{papillon2024topotune} has recently introduced a broad Generalized Combinatorial Complex Networks (GCCNs) framework that generalizes existing TNN models to date, and on top of which we will build ordered TDL. Let $\mathcal{C}$ be a combinatorial complex, $\mathcal{N_C}$ a collection of neighborhood functions, and $\mathcal{G}_{\mathcal{N}} = (\mathcal{C}_\mathcal{N}, \mathcal{E}_\mathcal{N})$ the strictly augmented Hasse graph for each neighborhood $\mathcal{N} \in \mathcal{N_\mathcal{C}}$. The $l$-th layer of a GCCN updates the embedding $\mathbf{h}^l_\sigma \in \mathbb{R}^{F_l}$ of cell $\sigma$ as
\begin{equation}\label{eq:gccn}
    h^{l+1}_{\sigma} = \phi \left(h^l_{\sigma}, \bigoplus_{\substack{\mathcal{N} \in \mathcal{N_C} , \sigma \in \mathcal{C}_\mathcal{N}}}\omega_\mathcal{N}\left(\{h^l_{\tau}\}_{\tau \in \mathcal{C}_\mathcal{N}},\mathcal{G}_{\mathcal{N}}\right)_{|_\sigma} \right) \in \mathbb{R}^{F_{l+1}},
\end{equation}
where $h_\sigma^0= h_\sigma$ accounts for the original cell feature vector; $\phi$ is a learnable update function; $\bigoplus$ is an inter-neighborhood aggregation module that synchronizes the possibly multiple neighborhood messages arriving on a single cell; and $\omega_\mathcal{N}:\mathbb{R}^{|\mathcal{C}_\mathcal{N}| \times F_l} \rightarrow \mathbb{R}^{|\mathcal{C}_\mathcal{N}| \times F_{l+1}}$ represents the \textit{neighborhood message function}, responsible of processing each Hasse graph $\mathcal{G}_{\mathcal{N}}$ and its corresponding embeddings --here we indicate by $\omega_\mathcal{N}(\cdot)_{|_\sigma}$ the corresponding output for cell $\sigma$.

%% file: sections/03-computer_networks.tex
Unlike traditional machine learning tasks that deal with structured data such as images or text, computer networks involve complex interactions between multiple components in a communication system. 
This section introduces key networking concepts and demonstrates how they can be reframed as instances of the fundamental building blocks of TDL.

\begin{figure}[!t]
    \centering
    \includegraphics[width=\columnwidth]{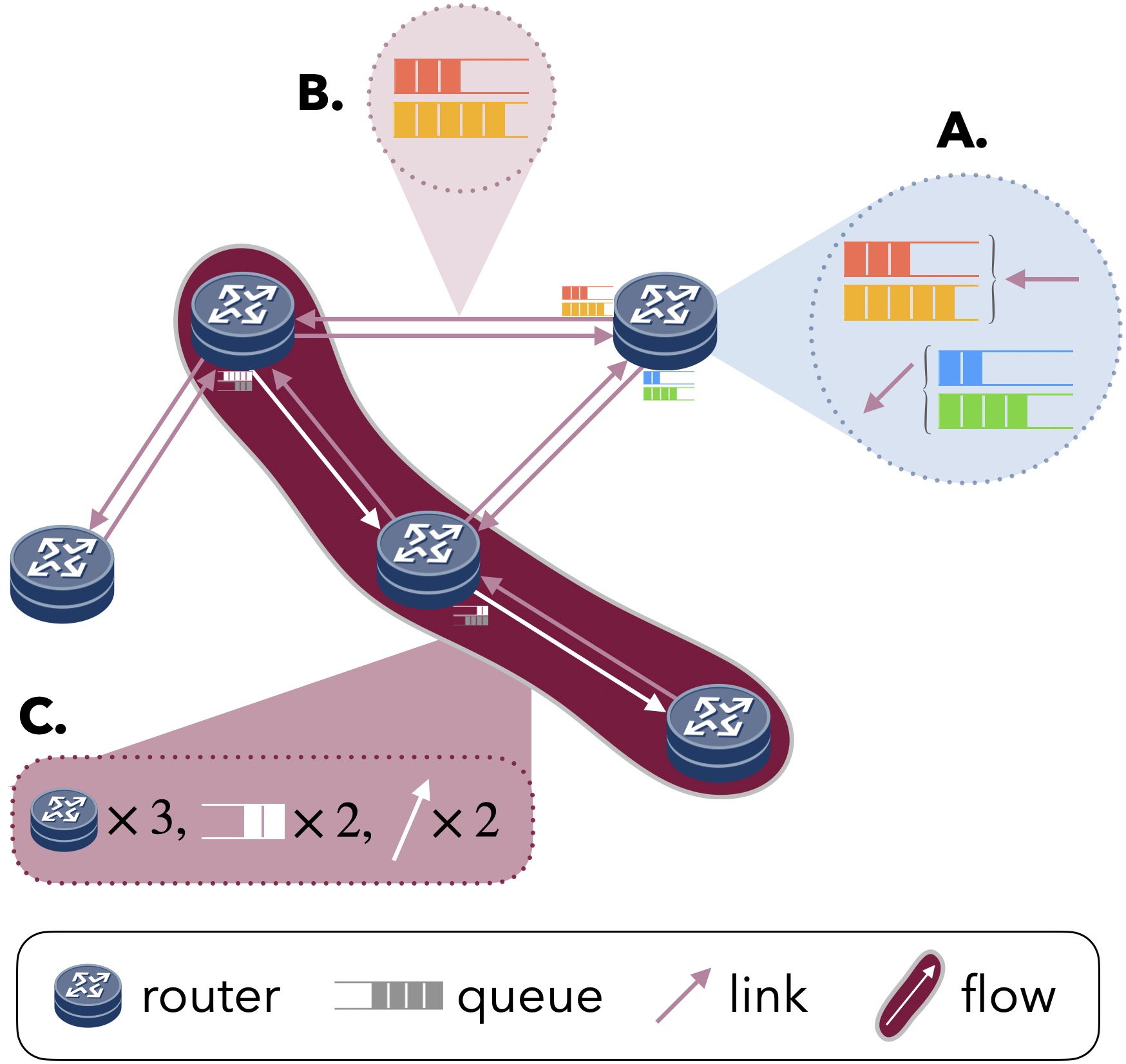}
    \caption{Topological representation of a computer network where relationships naturally form higher-order topological structures. A. Routers can be seen as a set of queues. B. Analogously, links can be devised as the collection of queues that inject traffic to them. C. Flow paths can admit several (combinatorial) interpretations, potentially encompassing the routers, queues and links they traverse.}
    \label{fig:computer_network}
\end{figure}

\paragraph{Basic Concepts.} In a nutshell, a computer network consists of the following components (illustrated in Figure~\ref{fig:computer_network}):
\begin{itemize}
    \item \textbf{Packets}: small chunks of data transmitted across the network.
    \item \textbf{Flows}: sequences of packets traveling from a specific source to a designated destination, representing a continuous stream of data between them.
    \item \textbf{Routers}: hardware devices that forward packets between different parts of the network.
    \item \textbf{Links}: physical or virtual connections between routers; its \textit{capacity} determines how fast packets can be transmitted.
    \item \textbf{Queues}: within a router, a queue is a list of data packets that wait to be forwarded through a particular link. 
\end{itemize}

\paragraph{Network Performance.} The \textbf{routing} defines how \textbf{traffic} (i.e., aggregated number of packets) is distributed across the network --determining the \textit{paths} of routers and links that packets follow  between any source-destination pair. Consequently, a good routing is of paramount importance to avoid \textit{network congestion}, and therefore can significantly impact key network performance metrics such as:
\begin{itemize}
    \item \textbf{Delay}: time taken for a packet to go from source to destination.
    \item \textbf{Throughput}: amount of data successfully transmitted per unit time.
    \item \textbf{Packet Loss}: percentage of packets dropped due to congestion.
    \item \textbf{Jitter}: The variation in packet delay, which impacts real-time applications like voice and video streaming.
\end{itemize}
An optimal network configuration would minimize delay, packet losses, and jitter, while maximizing throughput. 
Modern networks also require Quality of Service (QoS) guarantees to ensure that critical applications (e.g., video calls, cloud gaming) receive priority over less time-sensitive traffic. QoS is implemented through \textbf{scheduling policies}, which decide how packets are prioritized when multiple flows are assigned to different queues within a router for the same output link. 

\paragraph{Network Modeling.} Network modeling is a cornerstone of computer network research, design, and operations. 
By abstracting the network’s physical and logical components, it enables the simulation and analysis of network behavior, helping predict performance metrics (such as delay and jitter) under different configurations and workloads in a controlled environment.






\begin{tcolorbox}[
    colback=white,
    colframe=black,
    boxrule=0.5pt,
    left=5pt,
    right=5pt,
    top=5pt,
    bottom=5pt,
    boxsep=0pt,
    arc=10pt,
    outer arc=10pt
]
Mathematically, computer networks are commonly modeled as directed graphs $\mathcal{G}=(\mathcal{V},\mathcal{E})$, where the set of nodes $\mathcal{V}$ corresponds to hardware devices (like routers or switches), and the set of edges $\mathcal{E}$ represents the physical or wireless links that connect them. 
Nevertheless, alternative network abstractions can be reached by considering other logical elements of the network, such as queues and flow paths. 
In fact, and as illustrated in Figure \ref{fig:computer_network}, by analyzing the relationships between these components we can infer higher-order topological structures. 
Hence, \textbf{TDL naturally postulates as an underlying mechanism for modeling computer networks.}
\vspace{5pt}

However, TDL in its current form is not quite sufficient for this task. Specifically, some components of computer networks inherently induce an order over their elements --e.g. flow paths define sequences of queues and links when traveling towards its destination, or links internally assign priorities to its queues-- that existing TDL methodologies cannot account for. \textbf{This calls for a new form of TDL that is aware of order within cell neighborhoods.}


\end{tcolorbox}

%% file: sections/04-order_tdl.tex
\begin{figure*}[!t]
    \centering
    \includegraphics[width=0.95\linewidth]{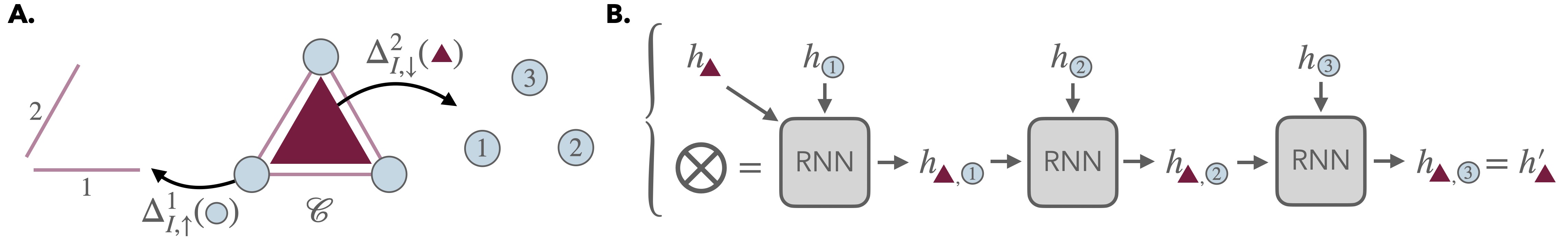}
    \caption{A. Example of an ordered combinatorial complex with two ordered cells: a 0-cell (node) defining an order in its two 1-cofaces, and a 2-cell inducing it on its three 0-faces. B. Considering the ordered 2-cell illustrated in subfigure A (right), this provides an example on how Recurrent Neural Network (RNN) based aggregators can exploit the sequence of ordered neighbors to get neighbor-dependent cell representations.}
    \label{fig:ordered_cell}
\end{figure*}

Upon reviewing the related literature and identifying the absence of TDL methodologies with neighbor ordering awareness, 
this section introduces and develops the fundamental concepts of \textit{ordered cells} and \textit{ordered neighbors} in combinatorial complexes --principles that apply to any discrete topological domain. Building on these foundations, we present the first general ordered TDL framework. 

\subsection{Related Works}
The role of order within cell neighborhoods is largely unexplored in TDL. Recent work in GNNs has explored incorporating an edge ordering during message aggregation on nodes to enhance expressive power. Unlike standard GNNs that use permutation-invariant functions (e.g., sum or mean) over neighbor messages, some approaches impose an order --either through sequential aggregators like LSTMs (as in \citet{hamilton2017graphsage}), or by explicitly structuring embeddings (as in \citet{song2023ordered}, which partitions features by hop distance). Other methods, such as port numbering techniques~\citep{garg2020generalization}, assign fixed local indices to neighbors, enabling the network to learn asymmetric functions over a node’s neighborhood. 

However, at the time of writing, no works within TDL leverage  neighbor-ordering approaches. Arguably, \citet{lecha2024higher} is the closest work, as it develops the related notion of higher-order topological directionality. In particular, this paper defines directed adjacencies among cells of a simplicial complex 
Lastly, \citet{cui2024hybrid} also introduces directionality to the hypergraph domain to tackle the challenging scenario of human pose prediction. Though related, the concept of directionality differs from the notion of ``order'' that we propose here. 

\subsection{Ordered TDL}

This section proposes a novel formalization of the notion of order in combinatorial complexes and TNNs.

\paragraph{Ordered Cells.} Let $\mathcal{C}$ be a combinatorial complex. A $k$-cell $\sigma\in\mathcal{C}$ is said to be \emph{ordered} if it induces an order $\leq_\sigma$ --either partial or total-- in any of its neighborhood sets $\mathcal{N}_{*}^r(\sigma)$: e.g., faces, cofaces, adjacencies. We denote by $\left(\mathcal{N}_{*}^r(\sigma), \leq_\sigma\right)$ the resulting ordered set. Analogously, we call $\mathcal{C}$ an \textit{ordered combinatorial complex} if it contains ordered cells.

\begin{remark*}
    Whenever $\leq_\sigma$ is not a total order, we assume it determines a unique maximal chain in the resulting partially ordered set $\left(\mathcal{N}_{*}^r(\sigma), \leq_\sigma\right)$ --that is, a unique totally ordered subset that cannot be extended by adding any other elements.
\end{remark*}

\paragraph{Ordered Neighbors.} 
Let $\sigma\in\mathcal{C}$ be an ordered $k$-cell, and $\mathcal{N}_{*}^r(\sigma)$ a particular neighborhood in which $\sigma$ infuses an order. We define the corresponding set of \textit{ordered neighbors} as
\begin{equation}
    \Delta_{*}^{r} (\sigma) := 
    \begin{cases}
        \left(\mathcal{N}_{*}^r(\sigma), \leq_\sigma\right), & \text{if totally ordered}, \\
        \text{maximal}\left(\left(\mathcal{N}_{*}^r(\sigma), \leq_\sigma\right)\right), & \text{if partially ordered}. \\
    \end{cases}
\end{equation}
Figure \ref{fig:ordered_cell}.A exemplifies this by visualizing the ordered 1-cofaces of a node (left), as well as the ordered 0-faces of a triangular 2-cell (right).
Notably, since $\Delta_{*}^{r} (\sigma)$ is always defined as a totally ordered set (toset), it uniquely defines an ascending order of its elements. For notational simplicity, we leverage this fact to denote the $i$-th element of the toset by $\Delta_{*}^{r} (\sigma)[i]$. Moreover, for any ordered neighbor $\tau\in \Delta_{*}^{r} (\sigma)$ we define the $\tau$-\emph{chain} as 
$$
\Delta_{*}^{r} (\sigma)[\leq \tau] := \Bigl( \{\, \delta \in \Delta_{*}^{r} (\sigma) \mid \delta \leq_\sigma \tau \,\}, \leq_\sigma \Bigr).
$$
In the remainder of the paper, we assume that whenever the elements of any such poset are iterated over, they are accessed in ascending order according to $\leq_\sigma$.

\paragraph{Sequential Aggregators \& Neighbor-Dependent Cell Representations.} Ordered cells offer a natural framework for introducing sequential aggregators into combinatorial complexes. By endowing a neighborhood $\mathcal{N}_{*}^r(\sigma)$ of a $k$-cell $\sigma$ with an order, the ordered neighbors $\Delta_{*}^r(\sigma)$ can be organized as a sequence rather than as an unordered set. In turn, this can be leveraged to induce "sequential" cell representations of the considered cell. More specifically, given a $k$-cell $\sigma \in \mathcal{C}$, we can generally define hidden embedding representations $h_{\sigma,\tau}$ for each ordered neighbor $\Delta_{*}^r(\sigma)$ as
$$h_{\sigma,\tau} = \bigotimes \left( h_\sigma , \left\{h_\delta\right\}_{\delta \in \Delta_{*}^r(\sigma)[\leq \tau]} \right) ,$$
with $\bigotimes$ a sequence-aware function. In this context, Recurrent Neural Network (RNN) models arise as a compelling alternative to traditional set-based aggregation methods, effectively capturing the dynamic, sequential relationships inherent in the complex --see Figure \ref{fig:ordered_cell}.B for a visual example with the ordered 0-faces of a triangular 2-cell.



\paragraph{Ordered GCCNs} By going beyond pure, unordered set-type relationships between the neighbors of a cell, the notion of \textit{ordered neighbors} represents a paradigm change in TNN architectures --the generality of GCCNs not being able to account for "sequential" updates. To address this, we propose a novel framework that extends GCCNs so as to formalize a wide range of ordered designs, which we call Ordered GCCN (OrdGCCN). Building upon the same setting as considered in Eq. \ref{eq:gccn}, let $\sigma \in \mathcal{C}$ be a $k$-cell, and $\tau \in \mathcal{N}^r_*$ designate an arbitrary $r$-rank ordered neighbor; the $l$-th layer of a OrdGCCN updates the embeddings of cell $\sigma$ as
\begin{equation}\label{eq:ordgccn1}
    h_{\sigma,\tau}^{l+1} = \bigotimes \left( h_\sigma^l , \left\{h^l_\delta\right\}_{\delta \in \Delta_{*}^r(\sigma)[\leq \tau]} \right) \in \mathbb{R}^{F^{l+1}},
\end{equation}
\begin{equation}\label{eq:ordgccn2}
\begin{split}
    h^{l+1}_{\sigma} = & \phi \bigg( h^l_{\sigma}, \bigotimes_{\tau \in \Delta_{*}^r(\sigma)} h_{\sigma,\tau}^{l+1},  \\
    & \bigoplus_{\substack{\mathcal{N} \in \mathcal{N_C} , \sigma \in \mathcal{C}_\mathcal{N}}}\omega_\mathcal{N}\left(\{h^l_{\tau}\}_{\tau \in \mathcal{C}_\mathcal{N}},\mathcal{G}_{\mathcal{N}}\right)_{|_\sigma} \bigg) \in \mathbb{R}^{F^{l+1}}.
\end{split}
\end{equation}
with $\phi$ an update function, $\bigoplus$ a permutation-invariant aggregator, and $\bigotimes$ representing order-aware aggregators.


\paragraph{Invariance} Neighbor ordering fundamentally breaks the permutation invariance inherent in traditional topological neural networks. These networks typically use symmetric aggregation functions like sum or mean, treating neighbors as an unordered multiset to preserve invariance. Introducing a fixed order in which neighbors update a cell’s feature makes the model sensitive to input arrangement, enabling a trade-off between invariance and expressivity. As the next subsection shows, breaking permutation symmetry can enhance the architecture’s expressivity.

\subsection{Expressivity of OrdGCCNs}\label{sec:expressivity}

In this subsection, we study the expressivity of OrdGCCNs with respect to their unordered counterparts. In particular, we generalize the combinatorial complex $k$-Weisfeiler Lehman ($k$-CCWL) test from \citet{papillon2024topotune} (Definition B.8) into an ordered version. This allows us to prove that OrdGCCNs are strictly more expressive than GCCNs. Specifically, we propose an ordered CCWL test which OrdGCCNs pass and GCCNs fail.

\begin{definition}[The Ordered CC Weisfeiler-Leman (Ord-CCWL) test on labeled combinatorial complexes]\label{def:ord-ccwl}
        Let $(\mathcal{C}, \ell)$ be a labeled combinatorial complex. Let $\mathcal{N}$ be a neighborhood on $\mathcal{C}$. The scheme proceeds as follows:
    \begin{itemize}[leftmargin=*]
        \item Initialization: Cells $\sigma$ are initialized with the labels given by $\ell$, i.e.: for all $\sigma \in \mathcal{C}$, we set: $c_{\sigma, \ell}^0 = \ell(\sigma)$.
        \item Refinement: Given colors of cells at iteration $t$, the refinement step computes the color of cell $\sigma$ at the next iteration $c_{\sigma, \ell}^{t+1}$ using a perfect HASH function as follows: 
            \begin{align*}
            c_\mathcal{N}^t(\sigma) &= \left( \left( c_{\sigma', \ell}^t ~|~ \forall \sigma' \in \mathcal{N}(\sigma) \right) \right),\\
            c_{\sigma, \ell}^{t+1}&=\operatorname{HASH}\left(c_{\sigma, \ell}^t, c_\mathcal{N}^t(\sigma)\right).
            \end{align*}
        \item Termination: The algorithm stops when an iteration leaves the coloring unchanged.
    \end{itemize}
\end{definition}

\begin{remark*}
    The only difference between the $k$-CCWL and the Ord-CCWL tests is the change from a multiset in the $k$-CCWL test to a tuple in the Ord-CCWL. Indeed, a multiset does not preserve the order in which the colors are gathered, while a tuple does. 
\end{remark*}

\begin{figure}[t!]
    \centering
    \includegraphics[width=0.85\columnwidth]{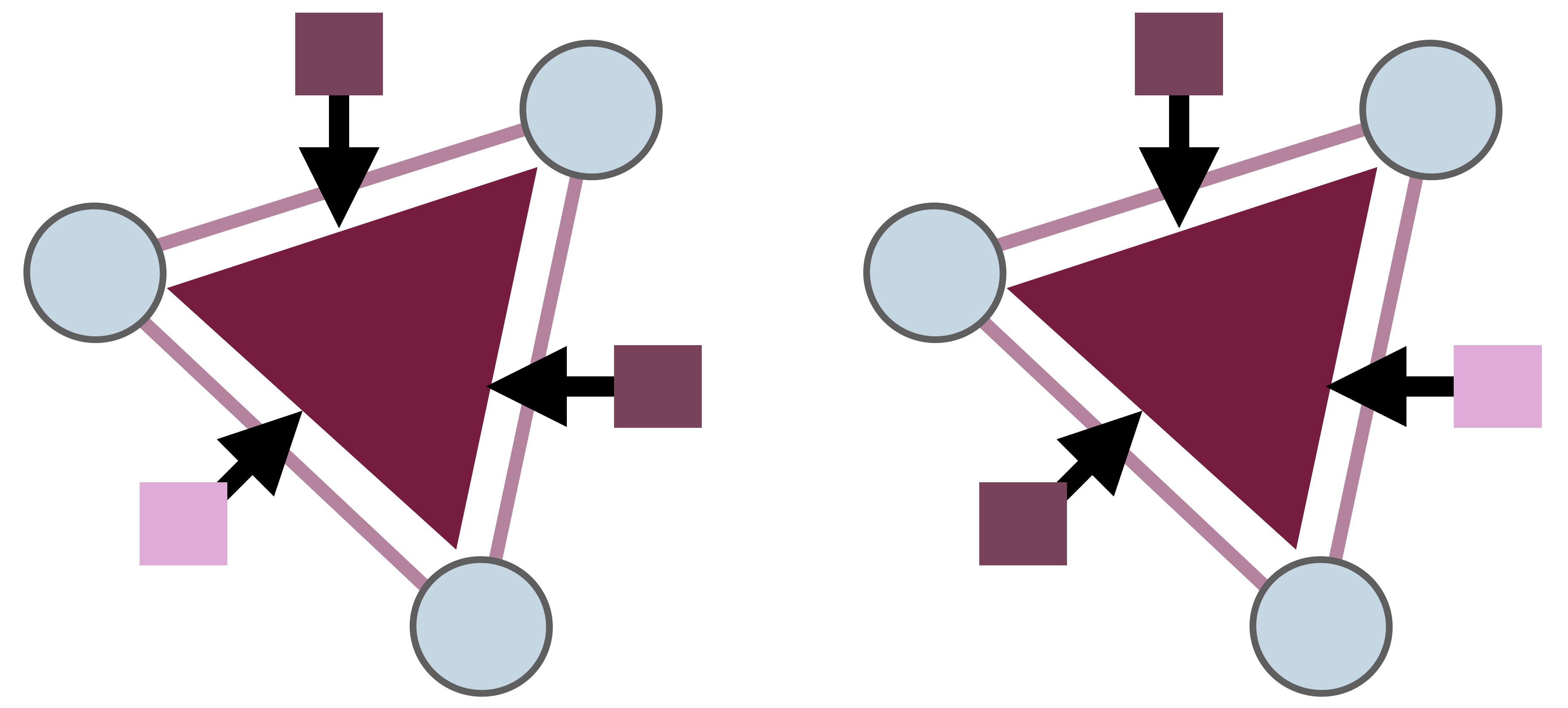}
    \caption{Two labeled simplicial complexes (a special type of combinatorial complex) that are indistinguishable by the $k$-CCWL test but distinguishable by Ord-CCWL. While both complexes share the same structure, they differ in edge labels (colored squares). Because $k$-CCWL is permutation invariant, it cannot identify the unique origin of the outlier label (light pink). In contrast, Ord-CCWL breaks permutation invariance, enabling it to distinguish between the two labeled complexes.}
    \label{fig:expressivity-proof}
\end{figure}

\begin{proposition}\label{prop:expressivity}
    Ord-CCWL is strictly more expressive than $k$-CCWLs.
\begin{proof}
    The proof exhibits a pair of combinatorial complexes that cannot be distinguished by $k$-CCWL but can be distinguished by Ord-CCWL -- shown in Figure~\ref{fig:expressivity-proof}. This result highlights that by allowing structured order in neighbor interactions, we can capture richer graph and topological features that unordered topological neural networks miss.
\end{proof}
\end{proposition}

Leveraging the previous result, we show that OrdGCCNs are more expressive than GCCNs --which, at the time of writing, are the most expressive TDL model.

\begin{proposition}\label{prop:expressivity2}
    OrdGCCNs are strictly more expressive than GCCNs.

\begin{proof}
    The proof relies on Proposition B.12 of \citet{papillon2024topotune}. This proposition shows that GCCNs are as powerful as the $k$-CCWL test. Since OrdGCCNs subsume GCCNs, we have that they are at least as powerful as the $k$-CCWL test. However, by allowing aggregators with neighbor ordering awareness, OrdGCCNs pass the Ord-CCWL test: they are able to preserve the order in which colors are gathered. In contrast, GCCNs' permutation invariant inter-neighborhood aggregator (and possibly their intra-neighborhood one, depending on the choice of neighborhood message function) prevents them from satisfying this test in the general case.
\end{proof}
\end{proposition}

%% file: sections/05-routenet.tex
RouteNet represents the state-of-the-art model in network modeling, achieving an unprecedented balance between performance and execution cost. By dissecting the two key design features of RouteNet architecture (i.e. network representation and architecture design), this section uncovers the intrinsic topological nature of these models, showing how the networking field has --unknowingly-- converged to TDL principles.


\begin{figure*}[!t]
    \centering
    \includegraphics[width=2\columnwidth]{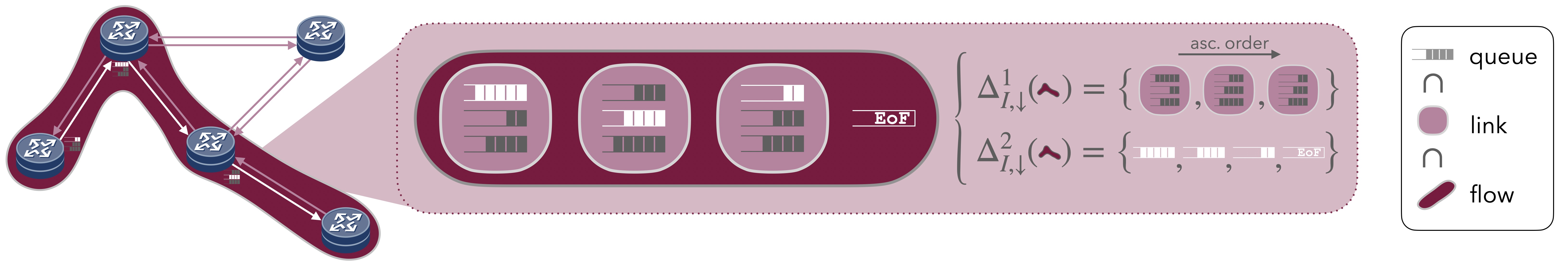}
    \caption{RouteNet's internal representation of a flow path as a set of links, depicted as an ordered combinatorial complex. Each flow link on the physical network (left, white arrows) is visualized as a 1-cell containing all queues that inject traffic into it (middle). The generic \texttt{EoF} queue simply marks the end of the flow.}
    \label{fig:flow_representation}
\end{figure*}

\subsection*{Network Representation}

\begin{tcolorbox}[
    colback=white,
    colframe=black,
    boxrule=0.5pt,
    left=5pt,
    right=5pt,
    top=5pt,
    bottom=5pt,
    boxsep=0pt,
    arc=10pt,
    outer arc=10pt
]
Drawing inspiration from QT, RouteNet operates over a network representation that goes beyond the straightforward graph structure made by routers and links. More specifically, it builds a heterogeneous graph that explicitly models the interdependencies between queues, links, and flows --each represented by a different type of node. The edges of such a graph connect each queue to the link in which they inject traffic, and each flow to every queue and link it traverses across its path.
\end{tcolorbox}

As a particular instance of Figure \ref{fig:computer_network}, we observe that this network representation exploits the hierarchical, topological-based relationships between the considered network elements. In fact, we show below that such a characterization admits a natural combinatorial complex representation: 
\begin{proposition} \label{prop:combinatorial_representation}
    RouteNet's internal representation of a computer network can be fully described as an ordered combinatorial complex $\mathcal{C}$ such that
    \begin{itemize}
        \item the set of nodes $\mathcal{V}$ represent all of the router queues;
        \item 1-cells $\mathcal{L}$ represent the wired links of the computer network, and they contain all the queues that inject traffic into them;
        \item 2-cells $\mathcal{F}$ represent the set of flows, and they contain all the link and queues they traverse. Moreover flows induce an order over both queues and links based on their path.
    \end{itemize}
\end{proposition}
\noindent Proof can be found in Appendix \ref{app:proof_combinatorial_representation}, and Figure \ref{fig:flow_representation} provides a visual example of how RouteNet internally interprets a flow path and a link from a topological perspective.

\subsection*{Message Passing Scheme}

\begin{tcolorbox}[
    colback=white,
    colframe=black,
    boxrule=0.5pt,
    left=5pt,
    right=5pt,
    top=5pt,
    bottom=5pt,
    boxsep=0pt,
    arc=10pt,
    outer arc=10pt
]
RouteNet architecture attempts to capture the hierarchical relationships and dependencies naturally occurring within its internal network representation, i.e.:
\begin{enumerate}[leftmargin=15pt]
    \item The state of flows (e.g., delay, throughput) is affected by the state of the queues and links they traverse (e.g., utilization), and taking into account the flow path order.
    \item The state of queues (e.g., utilization) depends on the state of the flows passing through them (e.g., traffic volume).
    \item The state of links (e.g., utilization) depends on the states of the queues that can potentially inject traffic into the link, and the applied queue scheduling policy. 
\end{enumerate}
To solve these circular dependencies, RouteNet leverages a heterogeneous GNN architecture with a three-stage message passing scheme that combines and updates the states of flows, queues, and links accordingly --more technical details in Appendix~\ref{app:routenet_technical_details}. 
\end{tcolorbox}

Notably, RouteNet's message passing procedure constantly exploits part-whole, incidence-based relationships between the elements, and it leverages the order induced by flow paths to aggregate queue and link states into flow representations. Hence, utilizing the order concepts and the OrdGCCN framework introduced in Section \ref{sec:order_TDL}, we finally show that RouteNet can be naturally described within TDL principles:

\begin{proposition} \label{prop:toporoutenet}
    RouteNet's internal modeling of a computer network can be formulated as an OrdGCCN (\ref{eq:ordgccn1}, \ref{eq:ordgccn2}). In particular, the hidden states' update equations for a flow $f\in \mathcal{F}$, a queue $q\in \mathcal{V}$, and a link $l\in \mathcal{L}$ are, respectively,
    \begin{equation} \label{eq:flow_update}
        h_{f}^{t+1} = \bigotimes_{q \in \Delta_{I,\downarrow}^2(f)} h_{f,q}^{t+1},
    \end{equation}
    \begin{equation} \label{eq:queue_update}
        h_{q}^{t+1} = \Phi \left(h_q^t, \bigoplus_{f \in \mathcal{N}_{I, \uparrow}^{2}(q)} h_{f,q}^{t+1} \right),
    \end{equation}
    \begin{equation} \label{eq:link_update}
        h_{l}^{t+1} = \Theta \left(h_l^t, \bigoplus_{q \in \mathcal{N}_{I, \downarrow}^{1}(l)} h_{q}^{t+1} \right),
    \end{equation}
    where 
    \begin{equation} \label{eq:face_dependent_update}
        h_{f,l}^{t+1} = h_{f,q}^{t+1} =\bigotimes \left( h_f^t , \left\{ \phi (h_{q'}^t,h_{l'}^t) \right\}_{\substack{ q' \in \Delta_{I,\downarrow}^2(f)[\leq q] \\  l' =  \mathcal{N}_{I, \uparrow}^{1}(q')}} \right)
    \end{equation}
    are the face-dependent flow representations.
\end{proposition}

\noindent Proof can be found in Appendix \ref{app:proof_toporoutenet}.
Notably, the aggregator $\bigotimes$ of Eq. \eqref{eq:face_dependent_update} is implemented as a RNN in order to exploit the order induced by flows over queues and links. Again, we refer to Appendix \ref{app:routenet_technical_details} for further technical details. 

%% file: sections/06-experiments.tex
\input{tables/simulated_real_world}
Having established that computer networks and RouteNet can be recast within (ordered) TDL, this section aims to better contextualize the roles that topology and order play in its success. 
To do so, we extend previous RouteNet evaluations by considering a new extensive benchmark of representative ML architectures --including different GNN models and RouteNet variants without ordered neighborhoods.

\paragraph{Experimental Setup.} Since previous works have already demonstrated RouteNet's superior performance in a wide variety of scenarios, we select for our evaluation a representative subset of configurations. To begin with, for the sake of simplicity we only focus on a single network performance metric: delay prediction (although RouteNet can also predict jitter and packet losses with analogous accuracy). We evaluate model performances across different network topologies, routing schemes, and traffic models, with datasets that include both simulated and real-world data. In all scenarios, we consider Mean Absolute Percentage Error (MAPE), Mean Absolute Error (MAE), and Mean Squared Error (MSE) as evaluation metrics for the predicted delay. More details about the training and evaluation datasets can be found in Appendix \ref{app:network_data}, while further experimental setup specifications are described in Appendix~\ref{app:routenet_experiments}.

\paragraph{Baselines.} For completeness, we consider traditional approaches (packet-level simulator~\citep{varga2001discrete}, QT-based network model) as baselines. However, unlike previous RouteNet studies, we enlarge the baseline suite with a diverse set of ML-based solutions --from traditional architectures (MLP, RNN) to well-known GNN models (GCN~\citep{kipf2017semi}, GAT~\citep{velickovic2017graph}, GIN~\citep{xu2018how}, MPNN~\cite{gilmer2017neural}). Moreover, we also test two novel variants of RouteNet: \textit{(i)} keeping the topological-based message passing, but removing order-awareness (GCCN); and \textit{(ii)} keeping both topology and order, but without leveraging a RNN (OrdGCCN). To ensure a fair comparison, we run all models in our evaluation --including RouteNet and the traditional approaches.

\paragraph{Varying Traffic Models.} Table \ref{tab:traffic_models} in Appendix \ref{app:networking_background} presents the performance of different models across various traffic patterns. The results show that traditional QT models struggle in complex traffic scenarios, as their rigid assumptions fail to capture real-world variability. In contrast, ML models consistently achieve lower error rates, some of them achieving a comparable performance to RouteNet.

\paragraph{Varying Routings and Topologies.} Still with simulated data, Table \ref{tab:simulated_real_world} shows how the performance of classical ML architectures significantly degrades when considering other routings or network topologies than those seen during training. In comparison, relational graph-based models better mitigate these variations --though RouteNet clearly outperforms them.

\paragraph{Real-World Data.} While some ML models perform relatively well in controlled settings, their accuracy significantly drops when tested on real-world data (due to unpredictable factors like bursty traffic, congestion, and routing changes). The same happens with packet-level simulators, whose idealized conditions do not fully capture the variability and complexity of real networks. RouteNet, however, maintains high accuracy across all conditions, demonstrating its superior adaptability to real-world network dynamics. 

\paragraph{The Role of Topology and Order.} Both Tables \ref{tab:traffic_models} and \ref{tab:simulated_real_world} shows the benefits of relational models (both graph- and topological-based) towards adapting to varying conditions. However, in the overall comparison we observe that the two considered RouteNet variants usually either match or improve GNN models' performance, suggesting that the hierarchical message passing plays a relevant role in modeling and capturing the network dynamics. Nonetheless, RouteNet still exhibits a significant performance gap relative to both GCCN and OrdGCCN variants --particularly on real-world traffic data--, underscoring the importance of fully exploiting flow-induced neighborhood orders through RNN-based aggregators.

%% file: tables/simulated_real_world.tex
\begin{table*}[!t]
\caption{Delay performance of different ML architectures for both simulated and real-world data under multiple routing configurations and topologies.}
\label{tab:simulated_real_world}
\centering
\resizebox{\textwidth}{!}{%
\begin{tabular}{ccccccccccccccccccc}
\toprule
     & \multicolumn{9}{c}{Simulated Data} & \multicolumn{9}{c}{Real-World Data} \\
     \cmidrule(lr){2-10} \cmidrule(lr){11-19}
     & \multicolumn{3}{c}{Same Routing} & \multicolumn{3}{c}{Different Routing} & \multicolumn{3}{c}{Different Topologies} & \multicolumn{3}{c}{Synthetic} & \multicolumn{3}{c}{Multi-Burst} & \multicolumn{3}{c}{MAWI~\cite{cho2000traffic}} \\
     \cmidrule(lr){2-4} \cmidrule(lr){5-7} \cmidrule(lr){8-10} \cmidrule(lr){11-13} \cmidrule(lr){14-16} \cmidrule(lr){17-19}
     & MAPE & MSE & MAE & MAPE & MSE & MAE & MAPE & MSE & MAE & MAPE & MSE & MAE & MAPE & MSE & MAE & MAPE & MSE & MAE \\
     \midrule
PL-Simulator & - & - & - & - & - & - & - & - & - & 53.68\% & 47.46 & 63.48 & 56.12\%  & 56.12\%  & 71.61 & 67.81 & 95.42 & 126.08 \\
\midrule
MLP & 12.3\% & 0.103 & 0.122 & 1150\% & 28.3 & 2.96 & 125\% & 3.69 & 1.03 & \(3.58\!\times\!10^{13}\)\% & \(4.95\!\times\!10^5\) & 418.60 & \(4.91\!\times\!10^{13}\)\% & \(1.45\!\times\!10^6\) & 649.36 & \(4.89\!\times\!10^{13}\)\% & \(4.57\!\times\!10^5\) & 261.69 \\
RNN & 10.0\% & 0.071 & 0.084 & 30.5\% & 0.553 & 0.282 & 63.8\% & 2.971 & 0.870 & 19.28\% & 1013.86 & 24.65 & \(2.79\!\times\!10^8\)\% & \(3.00\!\times\!10^{17}\) & \(2.54\!\times\!10^8\) & 29.98\% & \(1.21\!\times\!10^4\) & 49.52 \\ 
\midrule
GAT & 3.00\% & 0.051 & 0.041 & 15.00\% & 0.421 & 0.156 & 38.87\% & 1.923 & 0.589 & 23.52\% & \(2.30\!\times\!10^4\) & 28.15 & 29.02\% & \(1.48\!\times\!10^4\) & 32.29 & 29.96\% & \(1.17\!\times\!10^4\) & 48.12 \\
GCN & 2.93\% & 0.047 & 0.040 & 16.69\% & 0.456 & 0.172 & 36.90\% & 1.843 & 0.572 & 19.48\% & 847.08 & 23.29 & 19.08\% & 909.54 & 23.62 & 29.92\% & \(1.20\!\times\!10^4\) & 49.15 \\
GIN & 10.00\% & 0.203 & 0.098 & 16.10\% & 0.434 & 0.165 & 50.00\% & 2.435 & 0.718 & 19.78\% & 1148.26 & 25.98 & 531.77\% & \(7.35\!\times\!10^5\) & 544.56 & 30.04\% & \(1.16\!\times\!10^4\) & 47.85 \\
MPNN & 2.98\% & 0.049 & 0.040 & 15.28\% & 0.437 & 0.161 & 37.80\% & 1.873 & 0.576 & 19.48\% & 847.44 & 23.29 & 406.16\% & \(1.62\!\times\!10^6\) & 434.76 & 29.90\% & \(1.19\!\times\!10^4\) & 48.71 \\
\midrule
GCCN & 14.99\% & 0.249 & 0.234 & 6.67\% & 0.014 & 0.046 & 18.81\% & 0.167 & 0.147 & 19.27\% & 877.78 & 23.48 & 22.28\% & 1118.94 & 24.79 & 30.36\% & \(1.2\!\times\!10^4\) & 49.32 \\
OrdGCCN & 2.12\% & 0.005 & 0.025 &  1.69\% & 0.002 & 0.013 & 2.95\% & 0.004 & 0.027 & 92.77\% & \(1.7\!\times\!10^4\) & 110.44 & 90.43\% & \(1.7\!\times\!10^4\) & 110.22 & 13.35\% & 7659.53 & 26.47 \\
\midrule
RouteNet & \textbf{1.66\%} & \textbf{0.001} & \textbf{0.015} & \textbf{1.15\%} & \textbf{0.001} & \textbf{0.009} & \textbf{2.17\%} & \textbf{0.003} & \textbf{0.022} & \textbf{2.60\%} & \textbf{47.4655} & \textbf{3.1283}  & \textbf{2.28\%} & \textbf{71.6149} & \textbf{2.81}  & \textbf{14.75\%} & \textbf{95.4291}  & \textbf{7.22}   \\
\bottomrule
\end{tabular}%
}
\end{table*}

%% file: sections/07-implications.tex
This section highlights the main contributions and implications resulting from this work.

\paragraph{Advancing directed TDL through generality and expressivity.} 
By incorporating order-aware message-passing, OrdGCCN is the first model to explicitly accommodate new properties within TDL:
\begin{itemize}[leftmargin=15pt]
    \item order as a broader generalization of directionality in TDL;
    \item directionality, as a special case of order, within combinatorial complexes, rather than restricted to only simplicial complexes \cite{lecha2024higher} or hypergraphs \cite{cui2024hybrid};
    \item sequential aggregators, such as RNNs;
    \item neighbor-dependent cell representations.
\end{itemize}
\noindent Moreover, introducing order within TDL adorns OrdGCCNs with greater expressivity --as we show in Section \ref{sec:expressivity}--,  and might have important implications for addressing oversquashing and oversmoothing \cite{topping2022understanding}. This idea has been explored by \citet{song2023ordered} in ordered GNNs, showing that such models can maintain performance even when made very deep. This work opens the door to explore if OrdGCCNs could bring similar effects to TDL. 



\paragraph{Bridging the gap between TDL and real-world applications} 
To the best of our knowledge, this work studies the first state-of-the-art TDL application to a real-world setting, supported by extensive testbed experiments. By formulating RouteNet within the TDL paradigm using OrdGCCNs, we reveal a deeper theoretical structure underlying its success, and positioning network modeling as a naturally higher-order application domain. Moreover, our experiments shed some light on previous RouteNet evaluations, better contextualizing the relevance of exploiting both the topological relationships and the neighbor ordering.

\begin{figure}
    \centering
    \includegraphics[width=0.38\linewidth]{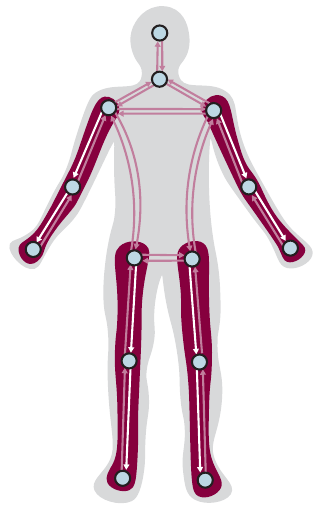}
    \caption{Skeletal representation of a human body as a combinatorial complex. Limbs that are correlated in movement are included in a higher-order cell (or, in \citet{cui2024hybrid}, a hyperedge). Order within this cell would represent flow of information from the ``root" nodes to the extremities of the body.}
    \label{fig:human}
\end{figure}

\paragraph{A unifying framework for ordered and directed applications.}
Historically, directed TDL methods have been developed either for synthetic applications \cite{lecha2024higher} or for specific real-world tasks without strong ties to the broader TDL community. For instance, directed hypergraph neural networks have recently emerged in human motion prediction \cite{cui2024hybrid}, yet these approaches remain isolated from the mainstream TDL and network modeling literatures. In Fig. \ref{fig:human}, we show how the same framework of Fig. \ref{fig:computer_network} naturally accommodates a skeletal representation of the human body with correlated limbs. We hope this methodology will help unify and accelerate advancements across diverse domains, laying the groundwork for novel directed/ordered TDL-driven applications.

%% file: sections/appendices/proofs.tex
\section{Proof of Proposition \ref{prop:combinatorial_representation}} \label{app:proof_combinatorial_representation}

\begin{proof}
Let $\mathcal{Q}$, $\mathcal{L}$ and $\mathcal{F}$ represent the queues, links and flows of the network, and consider the triplet $(\mathcal{V}, \mathcal{C}, \text{rk})$ where:
\begin{itemize}
    \item $\mathcal{V}$ is the set $\mathcal{Q}$ of all of the router queues across the network. For generality, we also include in this set a symbolic End of Flow (EoF) queue for each router.
    \item $\mathcal{C}$ is a set that contains
    \begin{itemize}
        \item all queue singletons $\{q\}$, $q \in \mathcal{V}$.
        \item All links $l\in\mathcal{L}$ of the network, considering each link as the set of queues that inject traffic to it.
        \item All flows $f\in\mathcal{F}$, considering each of them as the union of the links it traverses and the corresponding EoF queue.
    \end{itemize}
    \item rk is defined so that
    \begin{itemize}
        \item $\text{rk}(\{q\})=0$ for every queue singleton.
        \item $\text{rk}(l)=1$ for every link $l\in\mathcal{L}$.
        \item $\text{rk}(f)=2$ for every flow $f\in\mathcal{F}$.
    \end{itemize}
\end{itemize}
With this definition, we observe that:
\begin{enumerate}
    \item The set $\mathcal{C}$ is a subset of the powerset $\mathcal{P}(\mathcal{V}) \backslash\{\emptyset\}$, as each element $\sigma \in \mathcal{C}$ is a collection of queues.
    \item The rank function assigns to each element $\sigma in \mathcal{C}$ a nonnegative integer.
    \item Straightforwardly, for all $v \in \mathcal{V},\{v\} \in \mathcal{C}$ and $\textrm{rk}(\{v\})=0$.
    \item The rank function preserves order between queues, links and flows since:
    \begin{itemize}
        \item it is a valid assumption that links always contain two or more queues, so queues are always a proper subset of a link;
        \item by network principles, different links do not have queues in common, so multi-hop flows are larger subsets than each of its individual links;
        \item but even in one-hop flow scenarios, the link is always a proper subset of it: the flow also incorporates the final EoF queue that marks the end of the sequence, but the link doesn't.
    \end{itemize}
\end{enumerate}
Therefore, $(\mathcal{V}, \mathcal{C}, \text{rk})$ satisfies all the properties of a combinatorial complex.
Additionally, $(\mathcal{V}, \mathcal{C}, \text{rk})$ can be naturally expanded to an ordered combinatorial complex by considering the queue-link sequence defined by the path it follows through the network. In particular, each flow induces a total order in the set of its link faces (it traverses all of them), but a partial one in the set of its coincident queues (it only goes through one at each link) --as illustrated in Figure \ref{fig:flow_representation}.

\end{proof}

\section{Proof of Proposition \ref{prop:toporoutenet}} \label{app:proof_toporoutenet}

\begin{proof}
First, we show that the original RouteNet message-passing implementation (shown in Algorithm \ref{alg:rn_mp}) is a particular instance of Equations \eqref{eq:flow_update}-\eqref{eq:face_dependent_update}:
\begin{itemize}
    \item Flow neighbor-dependent representations: Equation \eqref{eq:face_dependent_update} summarizes lines 4-9 of Alg. \ref{alg:rn_mp} when $\bigotimes$ is implemented as a RNN.
    \item Flow update: Equation \eqref{eq:flow_update} expresses line 10 when $\bigotimes$ just selects the last element of the sequence.
    \item Queue Update: Equation \eqref{eq:queue_update} is equivalent to lines 12-16 when $\bigoplus$ is an element-wise sum.
    \item Link update: Equation \eqref{eq:link_update} matches lines 17-22 when $\Theta$ is a RNN and $\bigoplus$ simply gathers all incident queue hidden states; no particular order among the queues is considered in this module.
\end{itemize}

Lastly, we demonstrate that RouteNet update equations \eqref{eq:flow_update}-\eqref{eq:face_dependent_update} are particular instances of the OrdGCCN update equations \eqref{eq:ordgccn1} and \eqref{eq:ordgccn2}:
\begin{itemize}
    \item Flow neighbor-dependent representations: Equation \eqref{eq:face_dependent_update} follows the principles of the face-dependent update equation \eqref{eq:face_dependent_update}, slightly adapting it by grouping the well-defined queue-link pairs determined by the flow path (instead of inducing two different sequential representations for links and queues).
    \item Flow update: Equation \eqref{eq:flow_update} is an instance of Equation \eqref{eq:ordgccn2} where $\phi$ just selects the second argument.
    \item Queue update: Equation \eqref{eq:queue_update} is a special case of of Eq. \eqref{eq:ordgccn2} where $\phi$ disregards the third argument and $\bigotimes$ is a regular permutation invariant aggregator (so that we can write $\mathcal{N}_{I, \uparrow}^{2}(q)$ and $\Delta_{I, \uparrow}^{2}(q)$ indifferently).
    \item Link update: Equation \eqref{eq:link_update} can be obtained from Eq. \eqref{eq:ordgccn2} by considering only the first and last argument, and considering the 1-down incidence neighborhood of links.
\end{itemize}
\end{proof}

\section{Isomorphism of labeled combinatorial complexes}

We introduce the necessary background on homomorphisms and isomorphisms of combinatorial complexes.

\begin{definition}[CC-Homomorphism induced by $(\mathcal{N}_1, \mathcal{N}_2)$~\citep{papillon2024topotune}]\label{def:cc-n-hom}
    A homomorphism from a CC $\left(\mathcal{V}_1, \mathcal{C}_1, \mathrm{rk}_1\right)$ with neighborhood $\mathcal{N}_1$ to a CC $\left(\mathcal{V}_2, \mathcal{C}_2, \mathrm{rk}_2\right)$ with neighborhood $\mathcal{N}_2$, also called a CC-homomorphism induced by $(\mathcal{N}_1, \mathcal{N}_2)$, is a function $f: \mathcal{C}_1 \rightarrow \mathcal{C}_2$ that satisfies: If $\sigma, \tau \in \mathcal{C}_1$ are such that $\tau \in \mathcal{N}_1(\sigma)$, then $f(\tau)\in \mathcal{N}_2(f(\sigma))$. A labeled CC-homomorphism induced by $(\mathcal{N}_1, \mathcal{N}_2)$ is a CC-homomorphism induced by $(\mathcal{N}_1, \mathcal{N}_2)$ that additionally respects labeling of the cells, that is: if $\sigma, \tau \in \mathcal{C}_1$ have the same label, then $f(\sigma), f(\tau) \in \mathcal{C}_2$ also have the same label.
\end{definition}

This definition extends the definition of CC-Homomorphism from~\citep{hajij2023tdl} by being induced by neighborhoods.

\begin{definition}[CC-Isomorphism induced by $(\mathcal{N}_1, \mathcal{N}_2)$]
    A isomorphism from a CC $\left(\mathcal{V}_1, \mathcal{C}_1, \mathrm{rk}_1\right)$ with neighborhood $\mathcal{N}_1$ to a CC $\left(\mathcal{V}_2, \mathcal{C}_2, \mathrm{rk}_2\right)$ with neighborhood $\mathcal{N}_2$, also called a CC-isomorphism induced by $(\mathcal{N}_1, \mathcal{N}_2)$, is an invertible CC-homomorphism induced by $(\mathcal{N}_1, \mathcal{N}_2)$ whose inverse is a CC-isomorphism induced by $(\mathcal{N}_2, \mathcal{N}_1)$. A labeled CC-isomorphism induced by $(\mathcal{N}_1, \mathcal{N}_2)$ is a CC-isomorphism that additionally respects labels.
\end{definition}

%% file: sections/appendices/networking_background.tex
\input{tables/traffic_models}

This section revisits the most relevant methodologies currently leveraged in this network modeling field, as well as tours the process that lead to the emergence of RouteNet.

\paragraph{Packet-Level Simulators.}
Packet-level simulators, such as ns-3 \cite{riley2010ns} and OMNeT++ \cite{varga2001discrete}, provide detailed modeling by simulating network behavior at the granularity of individual packets. These tools offer high accuracy and flexibility, allowing researchers to include various features such as routing algorithms, scheduling policies, and traffic patterns. However, their main limitation is computational cost. Simulation time typically grows linearly with the number of packets processed, making these tools impractical for large-scale networks or real-time applications. 
Efforts to address this limitation have focused on improving efficiency through parallelization and machine learning \citep{gao2023dons,yang2022deepqueuenet,zhang2021mimicnet}, but --as illustrated in Figure \ref{fig:comparison_sota}-- they remain several orders of magnitude more computationally expensive than other alternatives.



\paragraph{Queuing Theory.}
Queuing Theory (QT) has been a basis of network modeling for decades and remains one of the most widely used analytical techniques in the field. Its importance stems from its ability to provide a mathematical framework to model networks as a set of interconnected queues. Each queue represents a network component, such as a router or switch, handling data packets. 
Unlike packet-level simulators, which require extensive computation to simulate individual packets, QT uses mathematical models to derive performance metrics directly. For example, it assumes that traffic follows a Poisson arrival process and that queues behave according to Markovian principles. These assumptions make QT computationally fast and well-suited for large-scale networks or scenarios where quick predictions are required. 
However, the strength of QT is also its main limitation: the simplifying assumptions that make it efficient often fail to capture the complexities of real-world network traffic. 
This is evident in the results shown in Table \ref{tab:traffic_models}, where QT's performance in delay prediction varies significantly across different traffic models.

\paragraph{Early Data-driven Approaches.}
Early applications of Deep Learning (DL) to network modeling explored architectures such as Multilayer Perceptrons (MLPs)~\cite{sadeghzadeh2008mlp, mestres2018understanding, wang2017machine} and Recurrent Neural Networks (RNNs)~\cite{belhaj2009modeling, mohammed2019machine, naseer2018enhanced}. These approaches leveraged the ability of neural networks to learn complex, non-linear relationships directly from data, bypassing the need for restrictive assumptions inherent in traditional methods like QT. However, while these models represented significant progress, Table \ref{tab:traffic_models} shows their inherent limitations. On the one hand, MLPs have a fixed structure, requiring predefined input sizes. This rigidity makes them less suited for network modeling, where topologies, configurations, and traffic patterns vary dynamically. RNNs, on the other hand, are designed to process sequential data, making them more adept at capturing temporal patterns in network traffic or packet flows. However, they are not inherently designed to represent the graph-structured inter-dependencies between flows, links, and queues --such as traffic on overlapping routes or shared network resources. 

\paragraph{Pairwise Relationships: The Emergence of GNNs.} The limitations of MLP and RNN models are further emphasized when network configurations deviate from those seen during training. As underlined in Table \ref{tab:simulated_real_world}, their performance deteriorates significantly with different routing configurations and changes in topology (e.g., link failures). In this context, GNNs emerged as a transformative tool for network modeling. By dynamically adapting their architecture to the structure of the input network graph topology, GNNs can effectively capture the relationships between different routers and links, thus enabling more accurate predictions of performance metrics across diverse network scenarios. Table \ref{tab:simulated_real_world} shows how different well-known GNN implementations (GCN~\citep{kipf2017semi}, GAT~\citep{velickovic2017graph}, GIN~\citep{xu2018how}, MPNN~\cite{gilmer2017neural}) clearly outperform previous ML architectures. However, despite the overall improvement, we can observe that vanilla GNN architectures still struggle to keep comparable performance when considering routing and topologies not seen in training.

\paragraph{Higher-Order Relationships: RouteNet.} The RouteNet family of models~\cite{rusek2020routenet, ferriol2022routenet, ferriol2023routenet, guemes2025routenet} represents a significant advance in this regard. Unlike vanilla applications of GNNs, RouteNet transforms the network's elements --such as flows, queues, and links-- into specialized representations that reflect their roles and interactions within the network. Inspired by principles from QT, RouteNet incorporates flow-level information and models the interdependencies between flows and network components (such as shared queues and overlapping routes). This transformation enables RouteNet to predict key performance metrics like delay, jitter, and packet loss with remarkable accuracy, even for unseen scenarios or topologies, while keeping an execution time comparable to that of QT methodologies --see Figure \ref{fig:comparison_sota}). Additionally, the subsequent versions of RouteNet have successfully overcome relevant real-world deployment challenges --e.g., by supporting several scheduling policies, scaling to networks way larger than those seen in training, or handling real-world non-stationary traffic. We further expand on these capabilities in Appendix \ref{app:evolution_of_routenet}.

%% file: tables/traffic_models.tex
\begin{table*}[!ht]
\caption{Delay prediction performance of QT and different ML architectures for different traffic models.}
\label{tab:traffic_models}
\centering
\resizebox{\textwidth}{!}{%
\begin{tabular}{ccccccccccccccccccc}
\toprule
     & \multicolumn{3}{c}{Poisson}  & \multicolumn{3}{c}{Deterministic}                 & \multicolumn{3}{c}{On-Off} & \multicolumn{3}{c}{A. Exponentials} & \multicolumn{3}{c}{M. Exponentials} & \multicolumn{3}{c}{Mixed}                \\
     \cmidrule(lr){2-4}
 \cmidrule(lr){5-7} \cmidrule(lr){8-10} \cmidrule(lr){11-13} \cmidrule(lr){14-16} \cmidrule(lr){17-19}
      & MAPE & MSE  & MAE  & MAPE & MSE  & MAE  & MAPE & MSE  & MAE & MAPE & MSE  & MAE & MAPE & MSE  & MAE & MAPE & MSE  & MAE \\
     \midrule
QT   & 12.6\%  & \textbf{0.001} & 0.017 & 22.4\% & 0.715 & 0.321 & 23.1\% & 0.784 & 0.363 & 21.1\% & 0.686 & 0.316  & 68.1\% & 1.10 & 0.798 & 35.1\% & 0.721 & 0.430  \\
\midrule
MLP  & 12.3\% & 0.103 & 0.122  & 23.9\% & 0.309 & 0.16 & 30.4\% & 0.438 & 0.240 & 84.5\% & 1.013 & 0.308 & 57.1\% & 1.058 & 0.363 & 41.2\% & 0.351 & 0.269 \\
RNN  &  10.0\% & 0.071 & 0.084 & 13.1\% & 0.083 & 0.070  & 15.2\% & 0.065 & 0.082 & 14.0\% & 0.070 & 0.072 & 57.8\% & 0.528 & 0.457 & 17.5\% & 0.036 & 0.080 \\
\midrule
GNN  & 2.98\% & 0.049 & 0.040 & 7.89\% & 0.214 & 0.098 & 10.79\% & 0.312 & 0.156 & 7.11\% & 0.156 & 0.078  & 8.00\% & 0.190 & 0.095 & 7.88\% & 0.207 & 0.103  \\
\midrule
GCCN & 5.77\% & 0.025 & 0.062 & 14.21\% & 0.149 & 0.120 & 14.99\% & 0.249 & 0.234 & 26.33\% & 0.996 & 0.472  & 17.82\% & 0.161 & 0.291 & 43.83\% & 1.321 & 0.598  \\
OrdGCCN & 2.12\% & 0.005 & 0.025 & 7.44\% & 0.123 & 0.090 & 7.89\% & 0.154 & 0.190 & 5.90\% & 0.060 & 0.104  & 30.08\% & 0.376 & 0.510 & 28.69\% & 0.881 & 0.479  \\
\midrule
RouteNet  & \textbf{1.66\%} & \textbf{0.001} & \textbf{0.015} & \textbf{0.71\%} & \textbf{0.007} & \textbf{0.005} & \textbf{0.79\%} & \textbf{0.008} & \textbf{0.006} & \textbf{2.88\%} & \textbf{0.059} & \textbf{0.030}  & \textbf{4.14\%} & \textbf{0.088} & \textbf{0.044} & \textbf{3.88\%} & \textbf{0.082} & \textbf{0.041} \\
\bottomrule
\end{tabular}%
}
\end{table*}

%% file: sections/appendices/evolution_of_routenet.tex
\paragraph{RouteNet: The Foundation}
The first version of RouteNet introduced a novel approach to network modeling by leveraging Graph Neural Networks (GNNs) to represent network structures. Traditional models, such as Queuing Theory (QT) and packet-level simulators, struggled with scalability and accuracy in complex network conditions. RouteNet addressed these limitations by modeling networks as graphs, where flows, links, and routers were represented as distinct entities with structured relationships. This formulation enabled RouteNet to learn complex network behaviors from data, providing accurate performance predictions for key metrics such as delay and packet loss.

\paragraph{RouteNet-E: Extending Generalization and Complexity}
While the first version of RouteNet introduced GNN-based network modeling, it operated under simplified assumptions. RouteNet-E extended its capabilities by incorporating Quality of Service (QoS), traffic models, and scalability—three essential aspects for real-world applicability. QoS is crucial in modern networks to prioritize critical traffic, ensuring low latency and high reliability for applications like video conferencing and cloud gaming. To model this, RouteNet-E introduced queue scheduling policies, including WFQ, DRR, and SP, allowing it to predict how different service classes impact network performance. Additionally, RouteNet-E addressed the limitations of basic traffic models by incorporating realistic traffic distributions, capturing bursty, autocorrelated, and heavy-tailed traffic patterns, making the model more aligned with real-world behavior. Another major advancement was scalability, a critical challenge for data-driven network models. Training such models requires datasets that include various traffic regimes, congestion levels, and failure scenarios, which are impractical to collect from production networks due to hardware constraints and operational risks. A common alternative is to use testbeds, but these are typically much smaller than real-world networks. RouteNet-E tackled this issue by generalizing across topology sizes, enabling the model to be trained on small-scale controlled testbeds while remaining accurate when deployed in much larger production networks. 

\paragraph{RouteNet-F: A Unified Model for Network Performance Prediction}
While RouteNet-E addressed individual challenges in network modeling, it lacked a unified mechanism to handle all of them simultaneously. RouteNet-F bridged this gap by integrating scalability, queue scheduling policies, and traffic models into a single framework. It was trained on diverse datasets that included various routing policies, congestion levels, and queuing mechanisms, allowing it to provide robust predictions across a wide range of network conditions. RouteNet-F was validated on networks up to 30 times larger than its training samples, showing minimal degradation in prediction accuracy.

\paragraph{RouteNet-G: Introducing Temporal Dynamics and Real Testbeds}
The latest iteration, RouteNet-G, introduced a key component by incorporating temporal dependencies. Unlike previous versions, which predicted performance metrics based on static snapshots of network states, RouteNet-G adopted a windowed processing approach. This allowed it to track how performance metrics such as delay and packet loss evolved over time, making it well-suited for networks with non-stationary traffic. Another critical advancement was its integration with real-world testbeds, enabling it to train on data collected from physical networks rather than relying solely on synthetic simulations. This shift significantly improved its ability to generalize to real-world deployments, bridging the gap between machine learning-based network modeling and practical network operations.

%% file: sections/appendices/routenet_technical_details.tex
In this appendix, we present the technical details behind the architecture of RouteNet models. While different versions of RouteNet introduce various enhancements, the general structure remains consistent. Here, we focus on the architecture of RouteNet-Fermi~\cite{10121482}, a representative model that captures the core design principles of the RouteNet family.  

The RouteNet architecture is inspired by Message Passing Neural Networks (MPNN)~\cite{gilmer2017neural}, a type of Graph Neural Network (GNN) designed to operate over graph-structured data. Like an MPNN, RouteNet can be summarized into four key phases: preparing the input graph, feature encoding, message passing, and readout.  

\subsection{Preparing the Input Graph}  
RouteNet models network traffic scenarios as a heterogeneous, directed graph $\mathcal{G} = ( \mathcal{N}, \mathcal{E})$. The set of nodes $\mathcal{N} = \mathcal{F} \cup \mathcal{L} \cup \mathcal{Q}$ consists of three key network components:  
\begin{itemize}
    \item $\mathcal{F}$: The set of traffic flows traversing the network, representing the transported data.
    \item $\mathcal{L}$: The set of links connecting network devices (e.g., routers, switches), forming the underlying topology.
    \item $\mathcal{Q}$: The set of queues within network devices, regulating the traffic before transmission. Each link \( l \in \mathcal{L} \) may have multiple queues assigned to it, depending on the scheduling policy. We define \( L_q(l) \) as the set of queues associated with link \( l \), meaning that these queues manage the traffic being transmitted through the link.
\end{itemize}  
The set of edges $\mathcal{E}$ defines the relationships between these components. These dependencies are extracted based on expert knowledge and summarized as follows:  
\begin{itemize}
    \item Flow performance (e.g., delay, throughput) is influenced by the state of the queues and links they traverse.
    \item Queue utilization depends on the volume and characteristics of the flows passing through them.
    \item Link utilization depends on the states of the queues injecting traffic into the link, as well as the applied scheduling policy (e.g., Strict Priority, Weighted Fair Queuing).
\end{itemize}  

\subsection{Initial Encoding of Features}  
Each node type (flow, queue, and link) is initialized with features relevant to its role in the network. These features include measurable parameters such as packet size, bandwidth, link capacity, and queue size. Continuous features are normalized using z-score normalization, and an MLP encoder maps them into a fixed-sized embedding specific to each network component type.

\subsection{Message Passing Phase}  
The message-passing phase extends the original MPNN framework to accommodate RouteNet's heterogeneous graph structure. In this iterative process, nodes exchange information with their neighbors to progressively enrich their embeddings with network-wide knowledge.  

At each iteration, every node undergoes the following steps:  
\begin{itemize}
    \item \textbf{Message Generation}: Each node creates a message containing relevant information for its neighbors.
    \item \textbf{Aggregation}: Each node collects messages from neighboring nodes according to an aggregation operator.
    \item \textbf{Update}: The node updates its embedding based on the aggregated information.
\end{itemize}  

The message and update functions are learnable and typically implemented as MLPs. The aggregation function is usually a summation (a commutative operation), though more structured operations such as RNNs can be used when the order of interactions is relevant, such as in flows traversing multiple links in a sequence.  

\begin{algorithm}[h] \caption{RouteNet's Message Passing Algorithm~\cite{10121482}} \label{alg:rn_mp}
\begin{algorithmic}[1]
\STATE {\bfseries Input:} $h^0_f, \forall f \in \mathcal{F}; h^0_q, \forall q \in \mathcal{Q}; h^0_l, \forall l \in \mathcal{L}$
\STATE {\bfseries Output:} $h^T_f, \forall f \in \mathcal{F}; h^T_q, \forall q \in \mathcal{Q}; h^T_l, \forall l \in \mathcal{L}$
\FOR{t=0 to T-1}
    \FORALL{$f \in \mathcal{F}$} \label{alg:rn_mp_flow_start}
        \STATE $\Theta([\cdot,\cdot]) \gets FRNN(\boldsymbol{h}^t_{f},[\cdot,\cdot])$ \hfill \COMMENT{FRNN Initialization}
        \FORALL{$(q,l) \in f$}
            \STATE $\boldsymbol{h}^{t+1}_{f,l} \gets \Theta([\boldsymbol{h}^t_q,\boldsymbol{h}^t_l])$ \hfill \COMMENT{Flow-Link: Aggr. and Update}
            \STATE $\widetilde{m}^{t+1}_{f,q} \gets \boldsymbol{h}^{t+1}_{f,l}$ \hfill \COMMENT{Flow-Link: Message Generation}
        \ENDFOR
        \STATE $\boldsymbol{h}^{t+1}_{f} \gets \boldsymbol{h}^{t+1}_{f,l}$ \hfill \COMMENT{Flow: Update}
    \ENDFOR \label{alg:rn_mp_flow_end}
    \FORALL{$q \in \mathcal{Q}$} \label{alg:rn_mp_queue_start}
        \STATE $M_q^{t+1} \gets \sum_{f \in Q_f(q)}  \widetilde{m}^{t+1}_{f,q}$ \hfill\COMMENT{Queue: Aggregation}
        \STATE $\boldsymbol{h}^{t+1}_q \gets U_q(\boldsymbol{h}^t_q,M_q^{t+1})$ \hfill\COMMENT{Queue: Update}
        \STATE $\widetilde{m}^{t+1}_{q} \gets \boldsymbol{h}^{t+1}_{q}$ \hfill\COMMENT{Queue: Message Generation}
    \ENDFOR \label{alg:rn_mp_queue_end}
    \FORALL{$l \in \mathcal{L}$} \label{alg:rn_mp_link_start}
        \STATE $\Psi(\cdot) \gets LRNN(\boldsymbol{h}^t_l,\cdot)$ \hfill\COMMENT{LRNN Initialization}
        \FORALL{$q \in L_q(l)$}
            \STATE ${h}^{t}_{l}  \gets \Psi(\widetilde{m}^{t+1}_{q})$ \hfill\COMMENT{Link: Aggregation and Update}
        \ENDFOR
        \STATE $\boldsymbol{h}^{t+1}_l \gets {h}^{t}_{l}$
    \ENDFOR \label{alg:rn_mp_link_end}
\ENDFOR
\end{algorithmic}
\end{algorithm}

The message passing phase is applied iteratively over $T$ steps. Flow updates are processed first, followed by queue updates, and finally, link updates. To improve efficiency, RouteNet-Fermi simplifies the message generation function using the identity function. The update function for queues, $U_q$, is implemented as an MLP, while the update functions for flows and links, $FRNN$ and $LRNN$, are implemented as RNNs. This ordering reflects the sequential nature of routing paths, allowing the model to maintain partial flow states that encode progressive network conditions.  

\subsection{Readout Phase}  
The final phase derives performance metrics from the enriched node embeddings. Each flow’s final performance is computed using its updated embedding after message passing. The method varies depending on the target metric:  
\begin{itemize}
    \item \textbf{Delay}: Computed by summing per-hop queuing and transmission delays. Queuing delay is estimated using a learnable function $R_{delay}$, while transmission delay is derived from flow bandwidth and link capacity.
    $$y_{f, delay} = \sum_{(q,l)\in f} R_{delay}(h^T_{f,l})/x_{l_c} + x_{f_{ps}}/x_{l_c}$$  
    \item \textbf{Jitter}: Computed similarly, using a learnable function $R_{jitter}$ that estimates variations in packet delay.
    $$y_{f, jitter} = \sum_{(q,l)\in f} R_{jitter}(h^T_{f,l})/x_{l_c}$$  
    \item \textbf{Packet Loss}: Obtained using a learnable function $R_{loss}$ that directly predicts the loss probability from the final flow state.
    $$y_{f, loss} = R_{loss}(h^T_f)$$  
\end{itemize}  

%% file: sections/appendices/network_data_details.tex
The data used to train, validate, and evaluate RouteNet is formed by a dataset of network scenarios. Each network scenario is composed of a computer network and its network traffic. From the network, we record its topology and device features --such as the link capacity or queue sizes. From the network traffic, we record the traffic matrix, routing paths, and performance metrics (delay, jitter, and packet loss), the latter being the target variables to predict. Scenarios are generated by randomly sampling the possible values for the four network variables: topology, traffic model, traffic intensity, and queuing configuration. Their details will depend on how the network scenarios are generated: through simulation or a real testbed network.

\subsection{Simulated Network Scenarios}
The OMNeTT++~v5.5.1~\cite{varga2001discrete} network simulator is used, specifically, the image found at \cite{github-bnnetsim}.

\paragraph{Network Topology} The topology can be one of the three most common topologies used in network research --NSFNET~\cite{hei2004wavelength}, GEANT \cite{barreto2012fast}, GBN~\cite{pedro2011performance} with 12, 24 and 17 nodes respectively-- or one of a set of scale-free synthetic topologies. These are generated using the Power-Law Out-Degree algorithm~\cite{palmer2000generating} to be similar in nature to identified real-world topologies --specifically those present in the Topology Zoo dataset~\cite{knight2011internet}-- and can reach up to 300 nodes.

\paragraph{Traffic Model} The temporal distribution of the traffic flows within the scenario --that is, the distribution defining the time spent between each packet's transmission-- follows one or all of the following options:
\begin{itemize}  
    \item \textbf{Poisson}: Packets are transmitted following a Poisson process, a commonly used model for network traffic that assumes arrivals occur randomly and independently over time. This model is widely used in network studies due to its analytical tractability, though real-world traffic often exhibits deviations from pure Poisson behavior.  

    \item \textbf{Deterministic}: Packets are sent at evenly spaced intervals to maintain a predetermined bit rate. This model is typical of real-time applications such as Voice over IP (VoIP) and video streaming, where consistent data delivery is required to avoid latency spikes and jitter.  

    \item \textbf{On-Off}: Traffic flows alternate between ON periods—where packets are generated following an exponential distribution—and OFF periods, where no packets are transmitted. The ON and OFF durations are parameterized to define the burstiness and traffic load. This model captures the behavior of applications that exhibit intermittent activity, such as web browsing and certain types of streaming traffic.  

    \item \textbf{Autocorrelated Exponentials}: This model extends traditional exponential inter-arrival distributions by incorporating temporal dependencies through an auto-regressive (AR) process~\cite{ferriol2022routenet}. Unlike memoryless Poisson arrivals, autocorrelated exponentials capture burstiness and short-term dependencies in traffic, better reflecting real-world packet arrival patterns observed in backbone networks. Parameters control both the level of autocorrelation and the shape of the underlying exponential distribution.  

    \item \textbf{Modulated Exponentials}: A hierarchical extension of autocorrelated exponentials, also introduced in~\cite{ferriol2022routenet}. This model incorporates a higher-level modulation process that varies the intensity of exponential packet arrivals over time, capturing more complex traffic fluctuations observed in large-scale networks. Such behavior is often seen in data center workloads, where traffic dynamics shift due to variations in application demand and scheduling policies.  
\end{itemize}  

\paragraph{Traffic Intensity} The traffic intensity within each scenario is randomly varied to capture a broad spectrum of network conditions, from low-utilization states to highly congested scenarios. However, the traffic load is controlled such that the packet loss rate never exceeds 3\%. 

\paragraph{Queuing Configuration}  
Each port in the network is configured with 1 to 3 queues, determining how packets are buffered before transmission. When a scenario includes multiple queues per port, flows are assigned a QoS class, which dictates which queue they should use based on priority levels.  

Queue sizes are set to 8, 16, 32, or 64 Kbits, affecting how much traffic can be buffered before packets are dropped due to congestion. The queuing configuration is defined by one of the following policies:  
\begin{itemize}  
    \item First In, First Out (FIFO) – Packets are processed in the order they arrive, without prioritization.  
    \item Strict Priority (SP) – Higher-priority queues are always served first, potentially starving lower-priority traffic.  
    \item Weighted Fair Queuing (WFQ) – Bandwidth is allocated among queues proportionally based on predefined weights, ensuring fairness.  
    \item Deficit Round Robin (DRR) – Similar to WFQ but optimized for efficiency in high-speed networks by dynamically adjusting transmission quotas.  
\end{itemize}  

Traffic flows are randomly assigned a Type of Service (ToS) value, which determines their priority level. Each router has a predefined configuration that maps ToS values to specific queues, ensuring that packets are forwarded according to the assigned service differentiation policy. This mechanism allows the simulation framework to model diverse traffic prioritization strategies, capturing the impact of QoS configurations on network performance.

\subsection{Captured Data from a Testbed}
To capture real traffic, an 8-router testbed is introduced in~\cite{guemes2025routenet}. The routers are interconnected through two high-capacity switches, forming a controlled but realistic experimental environment. The switches are also connected to traffic generator servers, which produce network flows for evaluation.  

Traffic is generated using the TREX software for synthetic traces and Tcpreplay for replicating real-world traffic traces stored in packet capture files. To ensure minimal overhead, traffic between the generators is optically copied to traffic capture servers, where key performance metrics such as delay, jitter, and packet loss are recorded.  

The testbed’s link capacities are selected to be representative of modern networks while ensuring that congestion only occurs at the routers. The router-to-switch links operate at 1 Gbps, while the server-to-switch connections run at 10 Gbps. The two core switches are interconnected with two 40 Gbps links in trunk mode, providing high-throughput capacity while preventing unintended congestion outside the routers.

\paragraph{Topology} Flow routing is defined through software, allowing for flexible topology configurations. Specifically, VLANs are used to virtually segment the network, enabling the testbed to replicate any 8-node or smaller topology. This approach provides a controlled environment where different network structures can be evaluated without requiring physical reconfiguration.  

In practice, 11 different topologies ranging from 5 to 8 nodes were used for the training and evaluation of RouteNet-Gauss~\cite{guemes2025routenet}. These variations allow the model to learn and generalize across different network layouts, ensuring robust performance when applied to real-world scenarios.

\paragraph{Traffic Model} Flows in a network scenario follow one of three traffic distributions. TREX provides two synthetic traffic distributions: synthetic traffic with high-frequency bursts and multi-burst traffic. A real traffic distribution was generated using the traffic patterns observed in real-world traces from the MAWI Working Group Traffic Archive~\cite{10.5555/1267724.1267775}. All flows within the same network scenario follow the same distribution, ensuring consistency in traffic behavior during evaluations.

\paragraph{Traffic Intensity} Traffic intensity varies randomly, never exceeding packet loss rates over 3\%. As a result, the number of packets generated per scenario ranges from 450 thousand to 16 million, ensuring a diverse set of network conditions for evaluation.

\paragraph{Queuing Configurations}  
The testbed supports a single, fixed-size queue per port, operating under a FIFO policy.

%% file: sections/appendices/routenet_experiments.tex
RouteNet has been evaluated in a variety of network scenarios, with experiments designed to progressively increase in complexity.  

\paragraph{Traffic model}  
Different versions of RouteNet were trained using simulated network scenarios in the NSFNET topology, keeping the routing fixed and using only a single traffic model during training. The model’s performance was then evaluated across all supported traffic models, including an experiment where multiple traffic distributions were present simultaneously.

\paragraph{Routing}  
RouteNet was trained and evaluated using simulated network scenarios in the NSFNET topology. However, during evaluation, routing configurations were unseen compared to those in training. Traffic models varied between samples.

\paragraph{Topology}  
RouteNet was trained with simulated network scenarios using the NSFNET and GEANT topologies and then evaluated on the GBN topology.

\paragraph{Scheduling Policies}  
The model was trained with simulated network scenarios in the NSFNET and GEANT topologies and then evaluated on the GBN topology with different scheduling policies.

\paragraph{Generalization to Unseen Topology Sizes}  
RouteNet was trained with synthetic topologies of up to 10 nodes, validated on topologies up to 50 nodes, and finally evaluated on networks ranging from 50 to 300 nodes. The training and evaluation samples included different routings, traffic models, and scheduling policies to assess the model's scalability.  

\paragraph{Real Traffic}  
RouteNet was trained on testbed data, incorporating both synthetic and real-world traffic distributions. The network scenarios varied in topology, routing, and traffic intensity. Due to testbed limitations, the same queuing policy was maintained across experiments. Each scenario lasted 5 seconds and was windowed into 100 ms intervals for evaluation.

%% file: sections/appendices/additional_benchmarks.tex
This appendix presents additional evaluation results of RouteNet that were not included in the main body of the paper. These results further demonstrate the model's effectiveness in handling different queuing policies and its ability to generalize to larger topologies.

\subsection{Impact of Scheduling Policies on Performance Prediction}

Table~\ref{tab:scheduling_delay} compares the performance of QT and RouteNet when modeling delay under different scheduling policies. The benchmark evaluates the models across low, medium, and high traffic intensities. While QT provides reasonable estimates in low-traffic scenarios, its accuracy deteriorates significantly as congestion increases. In contrast, RouteNet consistently outperforms QT across all traffic intensities, demonstrating its ability to capture the complex interactions introduced by scheduling mechanisms.

\begin{table}[!t]
\caption{Benchmark of QT and RouteNet in the presence of Scheduling Policies for low, medium, and high traffic intensity. Values show delay prediction performance.}
\label{tab:scheduling_delay}
\centering
\resizebox{0.6\columnwidth}{!}{%
\begin{tabular}{ccccccccc}
\toprule
     & \multicolumn{3}{c}{Traffic Intensity} \\
     \cmidrule(lr){2-4}
     & Low & Medium  & High  \\
     \midrule
QT & 13.0\% & 17.3\% & 25.1\%  \\
RouteNet & 0.80\% & 2.60\% & 7.31\% \\
\bottomrule
\end{tabular}%
}
\end{table}

\subsection{Generalization to Larger Topologies}

Table~\ref{tab:generalization_comparison} presents RouteNet's generalization performance when evaluated on networks significantly larger than those seen during training. The model was trained with topologies of up to 10 nodes and later tested on topologies ranging from 50 to 300 nodes. The results show that RouteNet maintains low error rates across different topology sizes, highlighting its scalability. This generalization capability is crucial for practical deployment, as training on large-scale real-world networks is often infeasible.

\begin{table}[!t]
\caption{Generalization results of RouteNet evaluated on increasing topology sizes. The model was trained with topologies of up to 10 nodes. Values show delay prediction performance.}
\label{tab:generalization_comparison}
\centering
\resizebox{0.75\columnwidth}{!}{%
\begin{tabular}{ccccc}
\toprule
Topology Size & MAPE & MSE & MAE  \\
\midrule
50  & 0.76\% & 0.00004 & 0.001  \\
100 & 1.89\% & 0.00016 & 0.003  \\
200 & 2.67\% & 0.00006 & 0.002  \\
300 & 2.45\% & 0.00003 & 0.001  \\
\bottomrule
\end{tabular}%
}
\end{table}

These results further support RouteNet’s robustness in handling complex network behaviors and its potential for deployment in large-scale environments.